\RequirePackage{silence}
\documentclass[%
	paper=A4,					
	twoside=true,				
	openright,					
	parskip=full,				
	chapterprefix=true,			
	11pt,						
	headings=normal,			
	bibliography=totoc,			
	listof=totoc,				
	titlepage=on,				
	captions=tableabove,		
	draft=false,				
]{scrreprt}%
\let\oldmathbf\mathbf
\usepackage{fourier}
\let\mathbf\oldmathbf

\newcommand{\thesisTitle}{From Dionysius Emerges Apollo}
\newcommand{\thesisSubTitle}{Learning Patterns and Abstractions from Perceptual Sequences}
\newcommand{\thesisName}{Shuchen Wu}
\newcommand{\thesisSubject}{Documentation}
\newcommand{\thesisDate}{May 22nd 2025}
\newcommand{\thesisVersion}{Final}

\newcommand{\thesisUniversityCity}{Tübingen}

\usepackage[utf8]{inputenc}		
\usepackage{amssymb}
\usepackage{pdfpages}

\usepackage{tabularx} 

\PassOptionsToPackage{pdftex}{graphicx}
\usepackage{graphicx} 
\usepackage{caption}
\usepackage{tocloft}      
\usepackage{subfig}  
\usepackage{epstopdf} 
\usepackage{comment}
\usepackage{pdfpages} 
\usepackage{nameref}
\usepackage{csquotes}
\usepackage{chemmacros}
\chemsetup[phases]{pos=sub}
\usepackage{animate}       
 
\usepackage{microtype}             
\usepackage{bibentry}
\usepackage[english]{babel} 
\usepackage[					
	figuresep=colon,%
	sansserif=false,%
	hangfigurecaption=false,%
	hangsection=true,%
	hangsubsection=true,%
	colorize=full,%
	colortheme=bluemagenta,%
	bibsys=biber,%
	bibfile=bib-refs,%
	bibstyle= numeric,%
    giveninits=false,%
]{cleanthesis}

\AtEveryBibitem{%
  \clearfield{note}%
  \clearfield{urldate}%
  \clearfield{month}%
  \clearfield{day}%
  \clearfield{isbn}%
  \clearfield{issn}%
  \clearfield{pmid}%
  \clearfield{pmcid}%
  \clearfield{eprint}%
}

\usepackage{hyperref}

\hypersetup{					
	pdftitle={\thesisTitle},	
	pdfsubject={\thesisSubject},
	pdfauthor={\thesisName},	
	plainpages=false,			
	colorlinks=false,			
	pdfborder={0 0 0},			
	breaklinks=true,			
	bookmarksnumbered=true,		%
	bookmarksopen=true			%
}

\begin{document}

\renewcaptionname{english}{\figurename}{Fig.}
\renewcaptionname{english}{\tablename}{Tab.}

\pagenumbering{roman}			
\pagestyle{empty}				

\begin{titlepage}

    \begin{center}

      \vspace*{\fill}

      \begingroup
        \emph{\huge \thesisTitle}
       \\ \bigskip 
      {\Huge{\color{black}{\thesisSubTitle}}}
      \medskip 
      \endgroup

      \vspace{7.pt}

      \begin{minipage}{0.8\textwidth}

        \begin{center}
          \Large{Dissertation}\\
          \vspace{7.5pt}
          zur Erlangung des Grades eines\\
          Doktors der Naturwissenschaften\\
        \end{center}

      \end{minipage}
      \\[2cm]


      {\Large der Mathematisch-Naturwissenschaftlichen Fakult\"at\\

        und\\

        der Medizinischen Fakult\"at\\[5pt]

        der Eberhard-Karls-Universit\"at T\"ubingen}
      \\[3cm]

      {\Large vorgelegt\\

        von\\[1cm]

        {\LARGE Shuchen Wu} \\[5pt]

        aus Jingdezhen, China}
      \\[1.6cm]
      {\LARGE 2025}
    \vspace*{\fill}
    \end{center} 
\end{titlepage}
\cleardoublepage
%
\pdfbookmark[0]{Declaration}{Declaration}
\chapter*{Declaration}
\label{sec:declaration}
\thispagestyle{empty}

Ich erkläre, dass ich die zur Promotion eingereichte Arbeit mit dem Titel: "Learning Patterns and Abstractions from Perceptual Sequences" 
selbständig verfasst, nur die angegebenen Quellen und Hilfsmittel benutzt und wörtlich oder inhaltlich übernommene Stellen als solche gekennzeichnet habe. Ich versichere an Eides statt, dass diese Angaben wahr sind und dass ich nichts verschwiegen habe. Mir ist bekannt, dass die falsche Abgabe einer Versicherung an Eides statt mit Freiheitsstrafe bis zu drei Jahren oder mit Geldstrafe bestraft wird.

I hereby declare that I have produced the work entitled "Learning Patterns and Abstractions from Perceptual Sequences", submitted for the award of a doctorate, on my own (without external help), have used only the sources and aids indicated and have marked passages included from other works, whether verbatim or in content, as such. I swear upon oath that these statements are true and that I have not concealed anything. I am aware that making a false declaration under oath is punishable by a term of imprisonment of up to three years or by a fine.
\bigskip

\noindent\textit{\thesisUniversityCity, \thesisDate}

\smallskip

\begin{flushright}
	\begin{minipage}{5cm}
		\rule{\textwidth}{1pt}
		\centering\thesisName
	\end{minipage}
\end{flushright}

\vspace*{\fill}

\begin{table}[ht!]
{\begin{tabular}{ll}

Tag der m\"undlichen Pr\"ufung: & 22/05/2025\\
&  \\
&  \\
Dekan der Math.-Nat. Fakult\"at: & Prof. Dr. Thilo Stehle \\
Dekan der Medizinischen Fakult\"at: & Prof. Dr. Bernd Pichler  \\
&  \\
&  \\
1. Berichterstatter:  & Prof. Dr. Eric Schulz  \\
2. Berichterstatter:  & Prof. Dr. Peter Dayan  \\
&  \\
   Pr\"ufungskommission:  & Prof. Dr. Eric Schulz   \\
  & Prof. Dr. Peter Dayan \\
  & Prof. Dr. Felix Wichmann \\
  & Prof. Dr. Georg Martius  \\
\end{tabular}}{}

\end{table}

\begin{titlepage}
	\pdfbookmark[0]{Cover}{Cover}
	\flushright
	\hfill
	\vfill
    \emph{\huge \thesisTitle\\}
    {\Huge \thesisSubTitle \par}
	\rule[10pt]{\textwidth}{1.2pt} \par
	{\LARGE\thesisName}
	\vfill
	\textit{\large\thesisDate} \\
	Version: \thesisVersion
\end{titlepage}

\newpage
\chaptermark{}
\mbox{}
\newpage

\hfill
\vfill
\chaptermark{}
\noindent{
\emph{“Reality is merely an illusion, albeit a very persistent one.”}\\[5pt]
\begin{flushright}
  \noindent {--- Albert Einstein}\\
\end{flushright}

\vfill

 
\pagestyle{plain}				
%
\pdfbookmark[0]{Abstract}{Abstract}
\chapter*{Summary}
\label{sec:abstract}
\vspace*{-10mm}
Immersed in chaotic and noisy sensory streams, we perceive a structured world. From learning stimulus-response pairings to grouping nearby visual parts, and from grammar acquisition to learning recurring sequential patterns, cognition irresistibly and swiftly breaks a high-dimensional sensory stream into familiar parts and gradually unveils their relations. Why do structures emerge, and how do they help us learn, generalize, and predict? What underlying computational principles give rise to this fundamental aspect of perception and intelligent behavior?

A sensory stream --- simplified to an extreme --- is a one-dimensional discrete sequence. In the process of learning such sequences, we naturally segment them into familiar parts — a well-known phenomenon called chunking.
In the first project, I investigated the factors that influence chunking behavior in a serial reaction time task. I showed that humans sensitively adapt to underlying chunks in sequences while exhibiting a resource-rational trade-off between speed and accuracy. 

Taking a step further, I built models that capture the chunk learning process on a computational and algorithmic level. The model learns chunks and parses sequences one chunk after another. From a normative standpoint, I proposed that chunking can be a rational way for an intelligent agent to discover recurring patterns and nested hierarchies in sequences, and, in turn, factorize sequences more effectively. I showed that sequential chunks can be learned as readily accessible primitives, ready for reuse, transfer, composition, and mental stimulation. This consequently allows the model to build up a complex understanding of the world by seeing the new via composing the known. 
I demonstrated and generalized this model's ability to learn hierarchies in both single and multi-dimensional sequence domains, and showed its applicability as an unsupervised pattern discovery algorithm. 

The second part of the investigation dives from the concrete domain to the abstract domain. I taxonomized two abstract sequence motifs and studied their implications for sequence memory recall. Behavioral evidence suggests that humans readily exploit redundancies in patterns in an abstract space for more efficient memory compression while transferring these motifs to novel sequences. 

Taking a step further, I propose a non-parametric hierarchical variable learning model that combines abstraction with chunking—abstracting those chunks that appear in similar contexts as variables while learning chunks also on a symbolic variable level --- gradually unearthing abstractions and discovering invariant patterns on a symbolic level. I demonstrate the algorithm's resemblance to human learning behavior and compare it with large language models. 

Taken together, this thesis suggests that chunking and abstraction as simple computational principles give rise to structured knowledge acquisition in sequences with underlying hierarchical structure, from simple to complex, from concrete to abstract, enabling the recursive construction of the highly complex from the ground up. I demonstrated the models' resemblance to human behavior and their algorithmic applications to discover patterns.   



\vspace*{20mm}

{\usekomafont{chapter}Zusammenfassung }\label{sec:abstract-diff} \\

Wenn wir in chaotische und laute Sinnesströme eintauchen, nehmen wir dennoch eine Welt mit fester Struktur wahr. Von der Zuordnung von Reiz-Reaktions-Paaren über das Gruppieren nahe beieinanderliegender visueller Elemente bis hin zum Grammatikerwerb und dem Erlernen wiederkehrender sequenzieller Muster: Die Kognition zerlegt unaufhaltsam und blitzschnell einen hochdimensionalen sensorischen Strom in vertraute Einheiten und deckt nach und nach deren Beziehungen auf. Aber warum entstehen diese Strukturen? Wie helfen sie uns beim Lernen, Verallgemeinern und Vorhersagen? Welches zugrundeliegende Berechnungsprinzip führt zu solch einer grundlegenden Komponente, die unser wertvolles Wahrnehmungs- und Intelligenzverhalten ermöglicht?

Ein sensorischer Strom kann stark vereinfacht als eindimensionale, diskrete Sequenz betrachtet werden. Beim Lernen solcher Sequenzen neigen wir dazu, sie in vertraute Teile zu unterteilen – eine tief verwurzelte Tendenz, die als Chunking bekannt ist.

Im ersten Projekt untersuchte ich, welche Faktoren das Chunking-Verhalten in einer seriellen Reaktionszeitaufgabe beeinflussen. Ich zeigte, dass Menschen sich flexibel an die zugrunde liegenden Chunks von Sequenzen anpassen und dabei einen ressourcenrationalen Kompromiss zwischen Geschwindigkeit und Genauigkeit eingehen.

In einem weiteren Schritt habe ich rechnerische und algorithmische Modelle entwickelt, die den Chunking-Prozess erfassen. Das Modell lernt Chunks und verarbeitet Sequenzen in diesen Einheiten. Aus normativer Perspektive schlug ich vor, dass Chunking für einen intelligenten Agenten rational ist, um wiederkehrende Muster und verschachtelte Hierarchien zu entdecken und so die Sequenzen effektiver zu faktorisieren. Ich zeigte, dass sequentielle Chunks als leicht zugängliche Primitive erlernt werden können, die zur Wiederverwendung, Übertragung, Komposition und mentalen Simulation bereitstehen. Dadurch kann das Modell ein tiefes Verständnis der Welt aufbauen, indem es das Neue durch die Kombination des Bekannten erschließt.

Ich demonstrierte die Fähigkeit des Modells, Hierarchien in ein- und mehrdimensionalen Sequenzen zu erlernen, und verallgemeinerte seine Anwendung als unüberwachter Algorithmus zur Mustererkennung.

Der zweite Teil der Untersuchung befasste sich mit dem Übergang von der konkreten zur abstrakten Domäne. Ich habe zwei abstrakte Sequenzmotive taxonomisch erfasst und ihre Auswirkungen auf das Gedächtnisretrieval untersucht. Das Verhalten zeigt, dass Menschen Redundanzen in Mustern im abstrakten Raum nutzen, um eine effizientere Gedächtniskompression zu erreichen, während sie diese abstrakten Motive auf neue Sequenzen übertragen.

Schließlich gehe ich noch einen Schritt weiter und stelle ein nicht-parametrisches, hierarchisches Modell zum Lernen von Variablen vor, das Chunking mit Abstraktion kombiniert. Hierbei werden Chunks, die in ähnlichen Kontexten wie Variablen auftreten, abstrahiert, während auch auf einer symbolischen Ebene gelernt wird, um schrittweise Abstraktionen und invariante Muster zu entdecken. Ich zeige, dass der Algorithmus dem menschlichen Lernverhalten ähnelt und vergleiche ihn mit großen Sprachmodellen.

Zusammenfassend zeigt diese Arbeit, dass Chunking und Abstraktion als grundlegende Rechenprinzipien zu einem strukturierten Wissenserwerb in Sequenzen mit zugrunde liegender hierarchischer Struktur führen. Vom Einfachen zum Komplexen, vom Konkreten zum Abstrakten wird so die rekursive Konstruktion hochkomplexer Strukturen von Grund auf ermöglicht. Ich habe nicht nur die Ähnlichkeit der Modelle mit dem menschlichen Verhalten aufgezeigt, sondern auch ihre algorithmischen Anwendungen zur Mustererkennung verdeutlicht.		
\cleardoublepage
\pdfbookmark[0]{Acknowledgement}{Acknowledgement}
\chapter*{Acknowledgement}
\label{sec:acknowledgement}
\vspace*{-10mm}

\cleanchapterquote{Every person is more than just oneself; s/he also represents the unique, the very special, and always significant and remarkable point at which the world's phenomena intersect, only once in this way, and never again. }{Hermann Hesse}

\\
\\
\\

I would like to express my deepest gratitude to the members of my supervisory committee, whose guidance and mentorship have shaped me into a better scientist throughout this journey. First and foremost, I would like to express my heartfelt gratitude to Eric, who has been a great mentor and offered unwavering support and advice throughout my PhD. He also demonstrated what it means to be a great PI and writer and taught me to see the bright sides. Eric’s wonderful personality to see the best in people and to welcome new collaborators and ideas have attracted outstanding individuals to the lab. It is especially encouraging to witness how the lab has transformed from a group of three people to a research institute. I would also like to express my deep gratitude to Peter, whose diligence and sense of responsibility have set a role model for scientific excellence. He has fostered an inclusive environment at the institute for many through pandemic or political hardships. Peter’s ability to attract brilliant minds to Tübingen has made precious opportunities to meet bright minds plentiful, which is instrumental to my growth. Felix, who has supported me since my master’s years, has been a constant source of encouragement. Felix was the one who opened the door for me to come to Tübingen, and for that, I will always be profoundly grateful.

I am indebted to excellent scientists such as Mirko and Susanne, who showed me the meticulousness of a psychologist in scrutinizing every detail of experimental design. I am also thankful to Ishita Dasgupta and Noemi Éltetö for helping me start my first projects. I have worked with wonderful students, including Mehmet Yörüten and Atilla Schreiber, who have taught me a great deal. I am thankful for colleagues who made the group a lively place of inspiring discussions and plenty of chuckles and ping-pong games, including Franziska Brändle, Akshay Jagadish, Tankred Saanum, Marcel Binz, Alex Kipnis, Tobias Ludwig, Julian Coda-Forno, Luca Schulze Bischoff, Kristin Witte, Xin Sui, Alicia Guzmán, and Can Demircan. 

I am grateful for having met a network of wonderful colleagues and friends from the institute and beyond, including Kaidi Shao, Judith Borowski, Robert Geirhos, Ju-Young Lee, Georgy Antorov, Yan Ma,  Weiyi Xiao,  Tianyuan Teng, Surabhi Nath, Sebastian Burjins, Gabriele Belluci, Franziska Bröker, Hanqi Zhou, Chuyu Yang, Wenting Wang, Qi Wang, Rui Tian, Charline Tessereau, Oleg Solopchuk, Tingke Shen, Daniel Shani, Lion Schulz, Turan Orujlu, Azadeh Nazemorroaya, David Nagy,  Kevin Lloyd, Ruiqi He, Junhao Liang, Chris Gagne, Sara Ershadmanesh, Florian Birk, Sahiti Chebolu, Stephan Bucher, Aenne Brielmann, Mihaly Banyai, Jiatong Liu, Jingyou Zou, Shervin Safarvi.
I am thankful for the labs of Stanislas Dehaene, Chris Summerfield, Timothy Behrens, Nikos Logothetis, Liping Wang, Tianming Yang, Zhang Peng, Anli Liu, Zhe Chen, and Zeynep Akata for hosting my visit during the PhD, and for exemplifying to me scientific works of utmost excellence, from which I have been truly inspired. I am thankful for Fernand Gobet's advice and reassurances during the discussion at restaurant Dionysius in Cogsci.

I owe a debt to the support of MPI staff: Kathrin Prax, Bianca Gäßer, Susan Fischer, Blake Fitch, Haydar Martin, and Joachim Werner. Joachim always saved me when I forgot to bring my laptop battery or needed server PHP debugging. Bianca and Kathrin are lifesavers, helping me tackle urgent paperwork in time.

I am deeply grateful to my experimental participants, who have contributed their time to the experiments. Their insightful suggestions for improving experimental design and even their one-time help with debugging went beyond what I could have anticipated. I was especially heartened by the encouraging feedback I received from participants expressing how much they enjoyed the experiments. Their responses have deepened my appreciation for both the data and the value of thoughtful user interface design, reinforcing my commitment to make the most of their contributions.

I am thankful for the anonymous reviewers of our papers. Their comments have reassured the work's contribution, and their suggestions have significantly improved this work.

Words cannot express my appreciation for the community at 1-West and Fichtehaus. Never would I expect myself to find anywhere else so close to feeling at home. Paula and Linda advocated for my move, believing I needed connections in Tübingen. I am deeply grateful for my flatmates, who have been similar to an extension of my family, including Linda, Ritika, Ruben, Lucia, Maxi, Jan, David, Charlotte, Juliane, Jo, Gregor, Kimo, Chiara, Luisa, Paula, Ximena, Malte, Flo, Chrisi, Steffi, Andy, Kristi, Christian, and Pascal, for the joys and sorrows that we shared, and the years that we grew together. Friends in the house, from whom I have been truly inspired through our discussions and experience organizing activities together, including Isa, Jacob, Anusha, Paul, Isabel, Nikos, Julius, Bea, Domi, Sophia, Pauline, Selena, Doro S., Doro R., Franziska, Sophie, Ruben, Felix, Roxana, Alexander, Jan-Lukas, Max, Lidewei, Praslav, and Parsival. I learned from human behavior and democratic decision-making to mastering Linsen Spätzle and hosting an unforgettable party. Thank you for making me feel constantly supported and acknowledged and for sharing a journey of growth with compassion. 

I am especially grateful to Qinhan for always listening to me and my parents' support throughout the years. My mom taught me to appreciate beauty and art, while my dad exemplified a spirit of embracing challenges and constantly seeking self-improvement. My grandma taught me to love people and be optimistic, and my grandpa taught me to be curious and try new things. Mimi Wu's unconditional love and affection as a tabby cat have been a constant source of comfort.

I am lucky to have met great minds whose ideas had long-lasting inspiration for this work before the start of this PhD. Kevin Martin for his advice on mindfulness; Matthew Cook for models of computation; Marin Osaki and  Lukas Vogelsang for many things, including floating along Aare, Chenxi Wu for variables; Lee Sharky for grounding startups; Aniruddh Galgali and Wenliang Li for London visits and Kai Sandbrink for being at almost every conference that I am also going to. 

Finally, I am grateful to Ralf Haefner for introducing me to embark on this journey from my undergraduate years. 
\cleardoublepage
\setcounter{tocdepth}{2}		
\tableofcontents				
\cleardoublepage

\pagenumbering{arabic}			
\setcounter{page}{1}			
\pagestyle{maincontentstyle} 	
\part{Introduction}\label{part:introduction}

%
\chapter{Introduction}
\label{sec:intro}

\cleanchapterquote{The propose of science is to find meaningful simplicity in the midst of disorderly complexity.}{Herbert Simon}{}

\label{sec:intro:motivation}
We effortlessly perceive entities from an immense volume of an ever-changing sensory stream. Consider the process of walking through a bustling city: successive sensations arise from a continuous influx of perception. You notice the swift changes in traffic lights, hear the diverse sounds of car horns, feel the differences of the surfaces you are walking on, and smell the fleeting aroma of a nearby food stand. Amidst this overwhelming flood of sensation, cognition skillfully sifts through the sensory stream, identifying concrete entities such as “traffic signals”, “road conditions”, and “food stand”. This remarkable capability allows us to swiftly navigate and interact with a complex environment.

Contrary to the ease of symbols emerging in our perception,  machine learning systems still struggle to learn structured symbols and factorization from data. Nowadays, large-scale artificial neural networks are trained on hundreds of thousands of graphical computing units with various optimization goals to auto-regress data from the internet \cite{lecun2015deep}. These models recently demonstrated remarkable performance in solving problems from recognizing speech and visual objects to playing chess \cite{lecun2015deep, bojarski2016endendlearningselfdriving,Mnih2015HumanlevelCT,silver2016}. Despite the triumph on the surface level, it has been argued that the resulting models still do not understand and represent observations as humans do \cite{saxton_analysing_2019,lake_generalization_2018}. Prominent models transform, predict, and generate sentences just by learning from lots of data and associating each part of the sentence with others \cite{zhang2021symbolic}, but this feature does not guarantee that entities of features shall reliably emerge within the model. Consequentially, large language models struggle to solve problems that involve symbolic manipulation, such as mathematical problems, which are never present in the training data \cite{zhang2021symbolic,rajani2019explain}; connectionist computer vision models still rely heavily on human-segmented scene data, hand denoted tags and human feedback to learn to segment images into visual entities \cite{geirhos2018imagenet,greff2020binding,zhang2020interpretable,battaglia2018relational}. Modern connectionist AI systems still struggle with the ability to distill object entities from data, and understand object permanence and the interactions and relations amongst these recurring entities --- an ability that simple animals are capable of \cite{shettleworth2010cognition,beran2014capuchin,spelke1990principles}.


The consequence of the gap between machine and human perception causes many problems. There are many problem domains that are trivial for humans but notoriously difficult for current connectionist machine-learning algorithms \cite{summerfield2020deep,sheahan2021structure}. The set of these problems includes --- but is not limited to --- continual learning \cite{summerfield2019perceptual,sheahan2021structure}, disentangling representations \cite{marcus2020next}, interpretability \cite{hill2019environmental, marcus2020next}, few-shot generalization \cite{chollet2019measure,hill2019environmental}, and abstractions \cite{sheahan2022neural,luyckx2019non}. These problems limit DNN's reliability in many safety-critical domains, such as the navigation of autonomous cars and medicine \cite{geirhos2020shortcut,hassabis2017neuroscience}. These problems involve precisely the compact, efficient, and reusable representation that comes to our cognition and perception easily. On the other hand, these characteristics are the pronounced characteristics of symbolic systems. As soon as thought can be expressed in a symbolic form, interpretability, generalization, reuse, and transfer can also be expressed in clearly articulated forms. 

Earlier approaches to studying artificial intelligence were predominantly symbolic. Herbert Simon, the father of modern artificial intelligence, posited that a system will only be capable of intelligent behavior if it operates and manipulates symbolic structures \cite{Simon1990, newellsimon1976}. AI in the last century was primarily about how precisely defined programs can manipulate symbols, such as automatically arranging and substituting mathematical or logical symbols to arrive at the next step of calculation or deduction \cite{winston1972artificial,mackay_problems_1982}, to solve equations \cite{mccarthy1960programs}, to parse languages using syntactical trees \cite{winograd1971procedures,bobrow1964natural}, or to partition images via visual grammar \cite{marr1982vision,zhu2021stochastic}. 
Symbolic models of AI represent computing entities as symbols and define operations between them, such as the definitions of constants and variables in mathematics, classifications of words into word types and morphological rules in natural language processing, or categorizations of image grammar in computer vision. This approach allows for the flexible reuse and recombination of symbols, which can be clearly implemented in programming languages across different domains. However, symbolic AI faces a fundamental challenge: most real-world data cannot be easily broken down into discrete parts that interact in straightforward ways. As a result, symbolic models have limited applicability and heavily depend on the programmer’s choice to design effective symbols and operational rules. 

A similar issue arises in probabilistic symbolic AI, where observations are modeled with probabilistic entities to account for noise and uncertainty. When an agent observes multiple sensory inputs that contain noise and makes decisions based on them, it needs to estimate a high-dimensional distribution, often represented as $P(x_1, x_2, \ldots, x_n)$, i.e., a complex global function with many variables. The distribution, by its raw form, is complex and infeasible to compute. However, this computing challenge can be dramatically alleviated if there are groups of random variables that are statistically independent of each other. In this case, then the high dimensional distribution can be factorized into subsets of variables such as $P(x_1,x_2)P(x_3,x_4)P(\ldots, x_n)$. The factorized form breaks the computation into smaller, independent parts, reducing both the time and complexity of calculations \cite{Kschischang2001FactorAlgorithm, Aji2000TheLaw}. This approach has been critical in probabilistic models like Markov random fields \cite{li1995markov}, Bayesian belief networks \cite{pearl1988probabilistic}, message-passing algorithms \cite{Kschischang2001FactorAlgorithm}, and it also applies to algorithms such as belief propagation \cite{yedidia2005constructing}, the Viterbi algorithm \cite{viterbi1967error}, the Kalman filter \cite{kalman1960new} and learning structural causal models \cite{pearl2009causality}. In computational neuroscience, researchers propose that the brain may similarly factorize high-dimensional sensory information into several independent modular systems, each handling a specific part of the computation in parallel, making use of the advantage of parallel computing neural systems inside the brain \cite{PIKOW2017943}. However, probabilistic AI faces challenges similar to symbolic AI: it is notoriously difficult to determine which groups of variables are statistically independent, and can be considered as within a factor to estimate the full distribution  \cite{DellabietraGreedyInducingFeatureRandomFields1997,SrebroNPhard2013,abbeel2012}.

How to come up with symbols and statistically independent groups of entities has been a challenge in symbolic and probabilistic AI. Symbolic AI systems depend heavily on human expertise to define the symbols and rules for manipulating them, while probabilistic AI struggles with the statistical complexity of determining the appropriate factorization without human input. At the same time, the missing perception of entities in connectionist AI contributes to the perception gap between humans and machines, causing their unreliability and lack of interpretability. These types of AI approaches do not have a clear answer to a fundamental feature that our cognitive system handles with ease: how do structured entities emerge in cognition? 

Patterns are fundamental to both our perception and actions. We easily recognize concrete entities like  “traffic signals”,  “road conditions”, and  “food stands” from a flood of sensory input, and we effortlessly perform action sequences, such as fetching a bottle, boiling water, and making tea. This question on where do cognitive entities come from --- intersecting artificial intelligence and cognitive science --- motivates the thesis to explore how cognitive entities emerge from perception, and their subsequent roles in factorization, interpretability, transfer and generalization, and how a computational model describing this process may bridge the gap between AI systems and human cognition.

The ability to perceive symbols and entities from an overwhelming and ever-changing sensory stream has historically been a topic of interest. William James famously described newborns' experience as a  “great blooming, buzzing confusion,” highlighting the immense challenge they face, immersing in noisy and fleeting sensory input to make sense of the world \cite{james1890principles}. Over time, the infants all learn to recognize coherent objects and develop sequences of actions to interact between these objects.

Philosophers have long speculated how the mind extracts meaningful patterns from sensory input. The British empiricists, in particular, emphasized that entities, including symbols and ideas, emerge from accumulating experiences from the sensory streams. John Locke proposed that newborns begin with a blank state of mind (tabula rasa) and that perceptual categories and entities form through experience \cite{locke1690}. David Hume described this process as beginning with vivid 'impressions' from perception, which later evolve into more abstract and less intense 'ideas' that are readily retrieved by the thinking process. The empiricists suggest that the mind gathers information and constructs knowledge in an additive manner: as interactions with the world accumulate, more complex ideas develop from earlier experiences \cite{hobbes1651, locke1690, hartley1749, mill1829, mill1843}. More complex mental structures can build on the previously learned ones \cite{hume1777}.

This idea was later taken up by behaviorist psychologists. Through studying animals and their behaviors, they proposed that animals establish behavioral structure by learning associations between stimuli via repeated practice and reinforcement \cite{pavlov1927}, \cite{thorndike1927, thorndike1928, thorndike1921, thorndike1922}. When stimuli repeatedly occur close together in time or space, the occurrence of one can evoke the memory of the other. Through practice, sequences of actions become associated and enforced \cite{guthrie1930, thorndike1913, skinner1953}, allowing animals to execute complex behavioral responses reliably.

Acquiring structured patterns from observation connects to our ability to transfer and generalize knowledge. Aristotle argued that structured perception is key to reasoning and inferring knowledge beyond our limited personal experience \cite{aristotle1938, AristotleAnalytics,  AristotleInterpretation, AristotlePosteriorAnalytics}. Similarly, Thorndike proposed that animals tend to reuse sequences of responses in new environments when these environments resemble situations they’ve previously encountered \cite{thorndike1913}. 

Before action and association, Gestalt psychologists proposed that the mind has an inherent tendency to organize perception into structure. In vision, for example, we tend to perceive nearby entities together as a whole, and farther entities as separate parts. Gestalt psychologists studied and characterized the tendency to organize complex sensory input into groups of parts, which are integrated into coherent wholes \cite{wertheimer1922,wertheimer1923}.
Through this grouping process, the mind identifies patterns and regularities and simplifies complex images, allowing perception to identify entities that can be related to prior knowledge. Today, the Gestalt grouping laws are still used as guiding principles of visual design to convey messages from abstract logos\cite{koffka1935}.

While language is considered by some a uniquely human ability \cite{dehaene_symbols_2022}, patterns and recurring entities are prominent in both concrete and abstract language levels. Concrete patterns manifested in words and phrases appear at the explicit level, and syntactical, morphological, and grammatical rules recur on an abstract level. Yet these rules still rigidly govern the composition of words into sentences. Understanding those rules is also critical to parse parsing sentences into coherent meanings. Mastering the application and usage of these patterns on concrete and abstract levels is critical for comprehending and using any language in flexible forms, allowing the infinite variations of meanings that language affords to convey \cite{Chomsky1965, Chomsky1980,fitch_artificial_2012}.

Empiricist philosophers, behavioral and gestalt psychologists, and linguists have historically considered the importance of perceptual entities and have related this feature with the powerful ability of humans to generalize and transfer. In the context of this thesis, to build on the previous observations and use a modern, programmatic approach to study this question, I set up this problem by studying sequences. This is because sequences are the most simplified form of sensory experience. Any sensory signal can be distilled into a series of successive perceptions and actions across different modalities. Discrete sequences are the most abstract simplification of perceptual data, preserving the core aspect of this problem while throwing away the unnecessary complexities. Processing sequences — whether through perceiving, memorizing, or retrieving temporally ordered elements — is fundamental to nearly all human activities, from recalling events and generating actions to using language and enjoying music.

This setup of the problem casts the question into how cognitive entities emerge from perceiving discrete sequences. In cognitive science, this process is characterized by \textit{chunking} \cite{Gobet2001ChunkingLearning, Gobet2004ChunksTwo,Dehaene2015TheTrees, Halford1998ProcessingPsychology}. 
As we parse a sequence such as “DFJKJKJKDFDFJKJKDFDF”, chunking refers to our tendency to segregate the sequence into a composition of recurring components, such as “DF”s and  “JK”s --- a behavior that even small babies and animals exert \cite{PERRUCHET1998246,Terrace1987ChunkingTask}. 

Chunking relates to memory organization \cite{Brady2009CompressionRepresentations,miller1956}. When we need to remember a sequence, we tend to break the sequence into several chunks. These chunks are then memorized and recalled as separate entities \cite{Mathy2012WhatsMemory,miller1956,Dempster1981MemoryDifferences,Bor2003EncodingDemand}. Chunks also serve as the units of our memory storage: we can hold between 4 to 7 chunks in our short-term memory. 
Therefore, knowing longer chunks by heart helps us to remember longer sequences. Imagine having to remember the sequence ``052917080461'': taken on its own, this sequence might be difficult to recall. However, knowing that it contains both John F. Kennedy's and Barack Obama's date of birth will likely simplify this task. Chunks can be subject to composition; we memorize sequences more easily by organizing them into a nested hierarchy of smaller chunks embedded in bigger chunks \cite{Planton2021}.

Apart from cognitive processing and memory organization, chunking also helps the organization of action sequences. We build up complex action sequence executions by piecing together familiar sequence chunks \cite{Verwey1996BufferKeypressing, Jin2014BasalSequences, Rosenbaum1983HierarchicalSequences, graybiel1998basal, Ellis1996SequencingOrder, Lashley1951TheBehavior}, fundamental to the process of planning \cite{Tomov2020DiscoveryPlanning}. Chunking perception, memory, and action into entities as basic units of cognitive processing are critical for language acquisition and usage \cite{PERRUCHET1998246,Planton2021, kaufeld_linguistic_2020}. Apart from action execution, memory, and sequence parsing, chunking has been observed in other sensory domains and has been theoretically linked to mental compression and optimal pattern discovery \cite{Schyns1998TheConcepts,Wickelgren1979ChunkingSystem, Egan1979ChunkingDrawings, Grainger2011AProcessing,Penhune2012ParallelLearning, Orban2008BayesianObservers, Brady2009CompressionRepresentations, Koch2000PatternsTasks}.   


I decided to study chunking in simplified sequences to examine our cognitive capability to learn structured representation from sequential perpetual data. Formally, the thesis defines sequences as made of discrete elements coming from a set of distinct symbolic items representing all the perceptual possibilities $\mathbb{A}$, which can be related to the unique sensory experience that an agent ventures. An example of such a sequence could be $010021002112000...$.
If an agent experiences such sequence on and on, a reflection of the world through this one-dimensional perception, recurring chunks amongst this signal may convey underlying entities in the world. 
For instance, the agent may perceive some consecutive occurrence of space-time observation as entities, such as learning identities in the sequence: $\{0,1,21,211,12,2112\}$, those identities will help the agent to parse the observation sequence one identity at a time. Hence, the model may use the biggest entity that it has learned about to partition the sequence:  \underline{0} \underline{1} \underline{0} \underline{0} \underline{21} \underline{0} \underline{0} \underline{2112} \underline{0} \underline{0} \underline{0}. 

This thesis constructs cognitive models based on previous knowledge about people’s sequences' learning capabilities. Critically, two components suggested by previous literature are included as a part of all models developed in this thesis. 

The first component assumes that the cognitive systems are capable of learning the associations between consecutively identified sequential units \cite{maheu_rational_2022, SAFFRAN199927,yang_universal_2004},  rooting back to the behaviorists' proposal associative learning is critical for learning complex behavior. This learning characteristic is supported by a rich set of evidence from statistical and associative learning literature \citep{chomsky.miller1958,Dulany1984,gomez_artificial_1999, SAFFRAN1996, gomez_variability_2002}. Examples include artificial grammar learning. Participants learn grammatical strings generated from a finite state language \citep{chomsky.miller1958} specified by a transition matrix defined on a set of artificial vocabulary. After exposure to the sequence, participants can judge a set of test strings' consistency with the language with above-chance accuracy \citep{Dulany1984}. Models that learn the associative transition probabilities between the sequential units can reproduce participants' performance in this task \citep{gomez_artificial_1999, SAFFRAN1996, gomez_variability_2002}. This thesis develops models with components that represent the occurrence frequencies of sequential entities and the transition probabilities between them.

The second component assumes that people process sequence into disjunctive chunks \cite{Orban2008BayesianObservers, miller1956}, resonating with proposals by Gestalt psychologists’ that perceptual ``parts'' are perceived together as ``wholes''. This learning characteristic is supported by the chunk learning literature \cite{Schyns1998TheConcepts,Wickelgren1979ChunkingSystem, Egan1979ChunkingDrawings, Grainger2011AProcessing,Penhune2012ParallelLearning, Orban2008BayesianObservers, Brady2009CompressionRepresentations, Koch2000PatternsTasks}. Examples include artificial grammar learning experiments, which showed that upon hearing continuous input streams made up of an artificial language containing underlying artificial words, children segmented the language stream into disjunctive recurring parts. This phenomenon can be explained by models that learn recurring disjunctive parts from sequences \cite{PERRUCHET1998246, anderson1990}. This thesis develops models that are inherent in this property of segmenting sequences into disjunctive chunks. 


The thesis proposes cognitive models on a computational and algorithmic level. Going beyond seeing chunking and statistical learning as heuristics or tendencies of human behavior, it takes a normative approach and hypothesizes that learning statistics and chunks are rational strategies adapted by a learning agent to discover underlying structures in sequences with embedded hierarchy. Combining the previous knowledge about human statistical learning and chunk learning, this thesis proposes that learning sequence statistics, when combined with learning chunks, become the seed for learning discrete segregated representations from discrete sequential data. In addition to learning cognitive entities, this thesis highlights a close relationship between the cognitive behavior of chunking and the topic of compositionality as a pressing problem puzzling both natural and artificial intelligence. 

Specifically, this question is addressed in two parts. The first part studies the emergence of chunks as segregated sequential patterns, and the second part studies the emergence of abstract entities in relation to chunks. For both parts, the questions are approached in conjunction with conducting cognitive experiments while exploring the algorithmic properties of models that exert such properties of computing principles.

The chapters are divided to address the four submitted projects that have resulted from this PhD work: 

\textbf{The emergence of chunks} 

\textbf{Chapter \ref{sec:srt}} starts from conducting behavioral experiments to study how action sequences can be segregated into parts: we manipulated the underlying statistical structure of the sequences and instruction demands in a serial reaction time task. We discovered that humans adapt their behavior to the statistics of the sequence and learn longer chunks are adaptive to the underlying chunk length in the sequence. Meanwhile, instruction focusing on speed versus accuracy also modulates human chunking behavior. We developed a computational model that learns chunks and optimizes a utility function that trades off between speed and accuracy to explain the population behavior observed in this task.  
\begin{itemize}
    \item “Chunking as a rational solution to the speed-accuracy trade-off in a serial reaction time task” (Wu, Éltető, Dasgupta, \& Schulz) \cite{wu_chunking_2023} was published in \textit{Nature Scientific Reports} 13, 7680 (2023), doi:10.1038/s41598-023-31500-3.
\end{itemize}


The second project (summarized in \textbf{Chapter \ref{sec:hcm}}) delves into the algorithmic implications of chunking under a normative lens. It hypothesizes a rational justification for the cognitive tendency to chunk, which is to uncover the underlying hierarchical structure in sequences. It designs a generative model for sequences with a nested hierarchical structure and develops an inverse recognition model HCM (hierarchical chunking model), which recursively reuses the previously learned representations to construct new complex composites. This project demonstrates chunking as a mechanism to discover recurring sequential primitives as entities for sequence factorization subject to compositionality, and also generalizes chunking from a one-dimensional sequential domain to a visual and visual temporal domain and applies the model as a data-driven structural discovery algorithm. This work resulted in the following publication: 

\begin{itemize}
\item “Learning Structure from the Ground up—Hierarchical Representation Learning by Chunking” (Wu, Elteto, Dasgupta, \& Schulz) \cite{wu_learning_2022} was published in \textit{Advances in Neural Information Processing Systems} 35, 36706 - 36721 (2022). 
\end{itemize}


The second half of the PhD work studies chunking and its role in learning abstraction in sequences. 

It starts with a sequence memory and recall experiment summarized in \textbf{Chapter \ref{sec:motif}}. The project taxonomizes two types of motifs and studies the influence of motifs in memorizing long sequences, in addition to transferring motif knowledge to novel sequences. The sequence recall experiment suggests that people exploit sequential patterns and redundancies not only on a concrete but also on an abstract level. Their learning and transfer behavior can be characterized by a motif learning model that chunks sequences on an abstract motif space. 

\begin{itemize}
    \item “Motif Learning Facilitates Sequence Memorization and Generalization” (Wu, Thalmann, \& Schulz) has been submitted as a preprint on \textit{PsyArXiv} \cite{wu_thalmann_schulz_2023} and has been accepted in \textit{Nature Communications Psychology}, doi:10.31234/osf.io/2a49z.
\end{itemize}

The experimental work inspired further modeling of human abstraction learning and its connection with learning by association and chunking. The last project summarized in \textbf{Chapter \ref{sec:hvm}} studies chunking in conjunction with abstraction under a normative perspective. A learning agent should learn abstract categories from observations because naturalistic sequences contain distinct objects of the same category types. Such properties enable the learning agent to learn recurring chunks from sequential data, layer by layer, starting from concrete and building up to more abstract levels. This model relates abstraction with compression and generalization while distinguishing large language models' learning and transfer behavior from human learning. 

\begin{itemize}
    \item “Building, Reusing, and Generalizing Abstract Representations from Concrete Sequences” (Wu, Thalmann, Dayan, Akata, \& Schulz) has been submitted as a preprint on \textit{ArXiv} \citep{wu_thalmann_schulz_2024} and, at the time of submission, is under review at \textit{International Conference on Learning Representations}, doi:10.48550/arXiv/2410.21332.
\end{itemize}

Some works are related to the thesis topic developed during the PhD but are not discussed in the thesis. This includes my work in the following two directions: 

One is to test the applicability of chunking principles in biologically realistic simulated neural circuits. I worked with Atilla Schreiber and Chenxi Wu from Giacomo Indiveri's group to demonstrate a group of spiking neurons with biologically plausible principles of synaptic plasticity and homeostasis can learn structured patterns from sequences in mixed-signal neuromorphic hardware. The project suggests that biological neural circuits and computation structures can learn chunks in computationally and power-efficient ways. 

\begin{itemize}
    \item “Biologically-plausible hierarchical chunking on mixed-signal neuromorphic hardware” (Schreiber, Wu, Wu, Indiveri, \& Schulz) was published in \textit{Machine Learning with New Compute Paradigms} workshop at  \textit{Advances in Neural Information Processing Systems} 36 \citep{schreiber2023biologicallyplausible}.
\end{itemize}

The other work studies the temporal dynamics of hierarchical visual grouping. I worked with Mehmet Yörüten to propose a psychophysics experiment studying people's recognition behavior upon seeing images tiled by amorphous sub-parts. We showed that the recognition difficulty of image parts can be described by a normalized min-cut algorithm that optimizes grouping by similarity under computational resource constraints.

\begin{itemize}
    \item “Normalized Cuts Characterize Visual Recognition Difficulty of Amorphous Image Sub-parts” (Wu, Yörüten, Wichmann, \& Schulz) was presented at the \textit{Computational and Systems Neuroscience (COSYNE)} conference \citep{wu_yoerueten_2024}.
\end{itemize}

Together, this thesis proposes how structured representation in chunks and abstract rules arises from learning from discrete sequences. The means to do so is via learning association to construct chunks and propose abstractions as cognitive entities, bringing forth computational efficiency and transferability. 

\part{From simple to complex}\label{part:chunking}

%
\chapter{Chunking in serial reaction time tasks}
\label{sec:srt}
\section{The emergence of action primitives from sequences}
\label{sec:srt:sec1}

A shared skill across learning, planning, problem-solving, and creativity is the capacity to break down complex tasks into manageable subcomponents, enabling humans to adapt and respond flexibly within complex, high-dimensional, and ever-changing environments \cite{tenenbaum_how_2011}.

One promising mechanism that underpins this ability is chunking. As a fundamental cognitive process, chunking facilitates the perception and execution of sequences. Lashley proposed that people organize complex actions by segmenting them into smaller, more manageable subsequences \cite{Lashley1951TheBehavior}. Through chunking, frequently occurring patterns within sequences are identified and grouped into discrete units that can be recognized as cohesive wholes \cite{miller1956, laird1984towards, graybiel1998basal, anderson1990}. This chunking phenomenon extends beyond sequence learning, playing a role in domains including grammar acquisition, visual and working memory tasks, function learning, and chess \cite{graybiel1998basal, Gobet2001ChunkingLearning, Egan1979ChunkingDrawings, Ellis1996SequencingOrder, Koch2000PatternsTasks, Chase&Simon1973}.

The ability to recognize and chunk recurring elements in sensory information allows for a compressed representation of long sequences \cite{Brady2009CompressionRepresentations}. These chunks can then be repurposed across different contexts, supporting the development of expertise as novices progress by building and recalling pattern-based chunks in long-term memory \cite{Mussgens2015TransferContext, Chase&Simon1973, Gobet&Simon1998}. Evidence suggests that the foundational units in cognitive hierarchies emerge through the process of learning and organizing chunks in sequences.

In the first study, we conduct cognitive experiments to study chunking in sequence learning. 
Since movements are underlying the most fundamental aspects of chunk execution, this work started by studying the emergence of units in sequences of movement execution. 
We want to study how chunks emerge from repeated exposure to simple units, so we chose a sequence learning paradigm.
A common paradigm to study sequence learning is the serial reaction time task (SRT) \cite{nissen1987attentional,willingham1989development,robertson2007serial,Koch2000PatternsTasks}. In SRTs, participants are instructed to press a number of keys that map to the displayed instruction cues. During the experiment, sequences of instruction cues appear consecutively on the screen; upon the occurrence of each instruction cue, participants react by pressing the corresponding key. 

In our experiments, participants were instructed to press four keys, denoted as A, B, C, and D, on the keyboard, which mapped to four instructions that appeared on the screen. If particular patterns, for example, ABC, keep repeating, then grouping repeated chunks as a unit should facilitate the prediction of the upcoming keys. Specifically, detecting a chunk's beginning, in this case, A, implies that the within-chunk items B and C will follow. This anticipation of the following elements of a given chunk can allow participants to anticipate what is coming next and thereby react faster \cite{Koch2000PatternsTasks,Mussgens2015TransferContext}. 

To study whether participants' chunking behavior adapts to task demands in an SRT task, we manipulated sequence statistics and instructions to participants during the training blocks to examine the behavior change from the baseline to the test block. By default, instruction sequences were generated from a non-deterministic, first-order Markovian transition matrix between the four instruction keys. Out of all 16 transitions specified between the four keys, the transitions from A to B and C to D were highly probable (P = 0.9), and the transitions from B to C and from D to A were medium probable (P = 0.7).  In this way, participants often observed reoccurring sequence segments such as AB and CD and could possibly perceive them as “illusory” chunks, even though the generative model was, as mentioned, nondeterministic first-order Markovian. In practice, the instruction keys were randomly mapped to D, F, J, and K for each individual to randomize the key correspondence to the keyboard placement of the fingers. 

\section{Summary of the article}
We propose a normative rational chunking model that learns chunks, taking two components into account. One is sequential statistics, i.e., sequences with distinct underlying chunks should result in different learned representations. By merging previously learned chunks, the model finds the best set of chunks to be learned when the entire sequence is considered, thereby segmenting the stream of symbols into compact units of chunks. The model learns patterns as chunks when there are underlying recurring patterns in the sequence. When reaction speed is preferred over accuracy, the model learns longer chunks while tolerating more mistakes. We test these predictions in two experiments, separately manipulating sequence statistics and instructions given. 

The first experiment examined how underlying chunks affect sequence learning, controlling for instructions. We gave participants first-order Markovian instruction during the baseline and the test block, sandwiching a training block, which we manipulated. During the training block, we trained three groups of participants separately on sequences containing no chunks (independent), chunk AB (size 2 chunk), or chunk ABC (size 3 chunk), and the elements that do not belong to the chunks occur randomly in the sequence. Consistent with the model's prediction, participants' reaction time data (comparing the test block with the baseline block to evaluate the influence of the training block) suggest that participants learn chunks when underlying chunks are in the sequence. 

The second experiment tested the prediction of the influence of instruction on chunking while controlling the sequences in all baseline, training, and test blocks to be the default Markovian sequences. In the training block, one group of participants was instructed and rewarded to perform the task as fast as possible, and the other group was as accurate as possible. The results of this second experiment suggest that the group focusing on speed chunked more than the group focusing on accuracy despite making more mistakes. The analysis result aligns with the model prediction that participants shall adapt their chunking behavior to optimize the trade-off between accuracy and speed. 

Our results shed new light on the benefits of chunking under specific task instructions and pave the way for future studies on structural inference in statistical learning domains.
\section{Discussion}

Our work can be related to several lines of previous research on chunking. Firstly, Servan-Schreiber and Anderson \cite{anderson1990} studied how chunking facilitates memory by examining subjects' memorization for artificially produced grammatical sentences. They proposed that a hierarchy of chunks forms as subjects remember sentences. They instructed subjects to memorize sentences chunked by distinct hierarchy levels (e.g., word level vs. phrase level) and examined subjects' judgment of grammaticality afterwards. Their result suggested that the hierarchy of chunks influenced participants' grammaticality judgments. Additionally, subjects overtly chunked the training sentences even when they were presented in an unstructured manner. These findings are similar to our current model, which also predicts chunking will appear in unstructured data but does so via a trade-off between accuracy and speed. The competitive chunking model provides a modeling framework consistent with our model but does not explain the processes that give rise to chunks' construction and hierarchy. The mechanism of recombining previously acquired chunks in our model can fill this blank.  

Another related model is PARSER. Proposed by Perruchet and Vinter \cite{PERRUCHET1998246}. PARSER can produce artificial language stream segmentations of continuous input streams without any episodic cues such as pauses \cite{SAFFRAN1996}. PARSER randomly samples the size of the next chunk of syllables and parses the sequence by disjunctive chunks. Each chunk the model learns is associated with a weight, which increments with observational frequency and decrements via a forgetting mechanism. Since both PARSER and our model evaluate chunks based on their occurrence statistics (PARSER approximates the chunk frequency online, whereas the rational chunking model evaluates the joint probability empirically), the simulation results produced by PARSER on syllable parsing can --in theory-- be reproduced by our model. Distinct from PARSER and unique to our model is the mechanism of conjoining acquired chunks to construct new chunks and relating the general chunking mechanism to a rational form of utility maximization.

Other methods use neural networks to learn chunks in sequences. Wang et al. trained a self-organized recurrent spiking neural network with spike-timing-dependent plasticity and homeostatic plasticity on sequences like the ones commonly used in SRTs and showed that it could reproduce several sequence learning effects, in particular, transfer effects \cite{Wang2017APlasticity}. It is, however, unclear whether the network learned explicit chunks that enabled this transfer because it is generally difficult to interpret the learned representations of such models. Compared to this approach, our model can serve as an interpretable computational level model because one can directly assess which chunks the model has learned. 

Another modeling approach to study how structure emerges from learning is to use variants of the Bayesian ideal observer framework \cite{Orban2008BayesianObservers,Goldwater2009,ODonnell2009FragmentLanguage}. These models are also rational because their inference is evaluated on the observational instances. The difference between these models and our model mainly lies in the context window and their structural assumptions. For example, with the hierarchical Dirichlet process model \cite{eltetHo2022tracking}, the maximal size of the context window, for recognition convenience, is pre-determined to evaluate the prediction of the next element given the previous context. In contrast, our model adapts its context length based on the previously acquired subsequences of chunks. Therefore, we think that these models are very similar to the accuracy part of our rational model and --in the limit-- might even make the same predictions for bigrammatic chunks. Apart from that, the rational chunking model accounts for the speed-accuracy trade-off, which is harder to realize and implement in a purely context-dependent Bayesian ideal observer framework relevant to the serial reaction time task.  

Finally, relying on the trade-off between speed and accuracy is one way that chunking benefits performance. Other mechanisms have also been proposed, such as minimizing memory or action complexity \cite{GERSHMAN2020104394}. Extending our current model to other domains using these additional complexity measures will make scalable predictions in memory, reinforcement learning, and planning.

\section{Limitation and future work}
This work has limitations. One is that chunk boundaries are inferred and non-explicit from reaction time speed up. We cannot say for sure when a participant formed a chunk that has been established in mind from analyzing reaction time alone. Future work can integrate the methods we used with other sources of information, such as eye movement data or measurements of brain activities, to cross-validate the estimation of human chunk boundaries. Alternative paradigms, such as asking participants to recall sequences freely, can also help to elucidate the demarcation of chunk boundaries in human behavior.

Our experiments examined learning from sequences with simple underlying chunks; future work may study learning sequences with more complex compositional structures within, which may adapt to participants' learning progress. 
For example, a model may infer participants' learning progress on the go and introduce novel chunk combinations, i.e., a concatenation of participants' previously learned chunks, up until a point when the participant has shown indicators of sufficient knowledge (presumably using the staircase method \cite{cornsweet1962}) in the two basic chunks to be composed. Such tasks may lead to an adaptive instruction sequence tailored to participants' idiosyncratic learning progress affected by individual tendencies of chunk building, consolidating, and concatenation. Modeling work may examine the influence of different chunk-building parameters on the dynamics of this adaptive learning process. Relating to real-life experience, this process may affect the progression of acquiring composable skills. It may explain phenomena such as the deeper participants are into a book, the faster the reading speed becomes, and the greater the size of a sentence being parsed \cite{rayner2006eye,ashby2005eye,schotter2011parafoveal}. In chess-playing, the model can simulate the progressive memory complexity reduction of strategic chess board configurations as a player advances from novice to expert level \cite{Gobet&Simon1998}. The implications of such findings may inform educational curriculum design to adapt to the learning progression of individuals. 

Finally, we have examined chunk learning in a simplified experimental setup in this project. Literature has suggested that such simple chunks can greatly benefit the composition of more complex action sequences  \cite{Schulz2017CompositionalLearning,schulz2020communicating, Tomov2020DiscoveryPlanning}, pointing to the direction that one primary consequence of chunk learning is to construct simple primitives which can be concatenated to complex composites \cite{tenenbaum_how_2011}. In the next project, we further dive into an algorithmic formulation of how simple chunks can recursively merge to create complex composites and a hypothesis on why chunk learning is rational for an agent to build up a better understanding of perceptual sequences via reusing the previously learned representations.


\section{Article Status}
”Chunking as a rational solution to the speed-accuracy trade-off in a serial reaction time task” (Wu, Éltető, Dasgupta, \& Schulz) \cite{wu_chunking_2023} was published in \textit{Nature Scientific Reports} 13, 7680 (2023), doi:10.1038/s41598-023-31500-3.

\section{Author Contributions}
\textbf{Conceptualization:} Shuchen Wu, Noémi Éltető, Ishita Dasgupta, Eric Schulz.\\
\textbf{Formal analysis:} Shuchen Wu, Eric Schulz.\\
\textbf{Software:} Shuchen Wu.\\
\textbf{Visualization:} Shuchen Wu, Noémi Éltető.\\
\textbf{Writing – original draft:} Shuchen Wu, Eric Schulz.\\
\textbf{Writing – review \& editing:} Shuchen Wu, Noémi Éltető, Ishita Dasgupta, Eric Schulz.
\chapter{Chunking to compose --- a source of infinite combination using finite means}
\label{sec:hcm}

\cleanchapterquote{Why does each new year seem to pass faster than the one before? \\
\medskip ... \\ We essentially conduct a lifelong process of chunking — taking small concepts and putting them together into bigger and bigger ones --- recursively building up a giant repertoire of concepts in mind.}{Douglas Hofstader}{}


Primary to Hofstadter’s speculation is a fundamental feature of chunking, which combines simpler components into more extensive and complex components that explain ever-larger recurring experiences in life. In the former project, the experiment on serial reaction time tasks suggests that humans adapt chunking behavior to the underlying regularities in the sequences. In this subsequent paper, I investigate the relation between chunking with hierarchical sequence structure, compositionality, and factorization. 

Cognitive scientists have suggested the vital role of chunking as a way to circumvent our inherent mental limitations. About half a century ago, Miller reported that our short-term memory is limited to holding 4 to 7 \textit{chunks} \cite{miller1956}. Once a chunk has been learned, it is memorized, identified, and parsed as a whole \cite{Egan1979ChunkingDrawings, Lashley1951TheBehavior, Verwey1996BufferKeypressing}. This discovery has led to ample subsequent work suggesting chunks as the primary information processing unit and serving a role in decomposing complex sequences into familiar parts. To illustrate, consider remembering a sequence like “schwarzwälderkirschtorte” --- a challenging task on its own --- a sequence with 26 items. However, knowing that it is a concatenation of the German words “Schwarzwälder” (Black Forest) and “Kirschtorte” (cherry cake) simplifies the task, as the sequence is decomposed into several familiar parts. Recognizing familiar subsequences aids in remembering more complex sequences. 

The process of breaking perception into several entities goes beyond sequences to the visual domain: Gestalt psychologists have observed and developed the notion of ‘Prägnanz’: upon processing a complex and chaotic visual scene, people tend to organize and group visual perceptual units together into coherent wholes \cite{wertheimer1922,wertheimer1923}. A primary tendency to group perceptual units as a whole is by proximity: entities close to each other tend to be perceived together to form a group \cite{wertheimer1922,wertheimer1923}. Interestingly, the grouping by proximity in vision resembles the grouping by chunks in sequences, suggesting the chunking principle as a candidate to break complex observation into parts in both visual and sequential domains.

If chunking can be a candidate of a cognitive principle that underlies many domains, what could be a normative reason to justify the rationality of learning chunks? From a computational standpoint, upon processing streams of perceptual sequences, a learning agent faces an inherent challenge in learning structure from the sequence for better predictability, memorization, and recall upon task demand. From this point of view, chunking implies several attractive algorithmic features. 

The first is that chunks can serve as computational processing entities to explain the emergence of symbols. This points to a potential answer to the unresolved problem in symbolic AI. Symbolic AI systems study the consequence of intelligent behavior via operating with symbols but do not address where symbols come from (sometimes, they resort to some innate explanation that circumvents this problem). Because symbolic AI relied on this fundamental assumption, their approach suffered difficulty in scaling up to higher dimensions, as going into any more complex data domains will reveal the problem of finding primitive symbols to parse the data reasonably. The formation of chunks through learning offers a potential answer to how symbols and distinct entities emerge from experience. By enabling us to segment observations into discrete components, chunks serve as foundational units that transform input sequences into recognizable parts.
 
The second is that chunks can become independent entities to factorize a probabilistic estimation of the observational sequence. An observational sequence with $n$ entities can be described as distributed in a high dimensional probability distribution $P(x_1, x_2, …, x_n)$. Chunks become a unit of sequence parsing, and thereby, a sequence spanning in many observational units can be partitioned by grouping observational units as chunks, each chunk occurring independently from the occurrence of other chunks, and therefore factorizing the observational sequence distribution by chunks of consecutively occurring stimuli: $P(x_1,x_2)P(x_3,x_4)P(…, x_n)$. This suggests that the mechanism of learning chunks may help computer scientists find ways of circumventing the computational complexity of factorizing high-dimensional distribution of observational sequences and can also serve as a way of learning a generative model of the sequences. 

What follows as the third attractive algorithmic property is compositionality: the previously learned chunks can be composed into more complex chunk combinations. Previous work using symbolic systems to study human behavior has suggested that the composition of primitive operations explains both human behaviors of learning a hierarchical organization of the primitives \cite{Lake2015Human-levelInduction}, which helps people to make fast structure learning and generalization \cite{Schulz2017CompositionalLearning}. Additionally, the components that are composed may contribute to the processes of generalization and transfer between separately learned sequences \cite{Mussgens2015TransferContext}. Chunk learning and chunk composition may also lead to chunk transfer. 

\section{Related Work}
Several approaches to model chunk learning from sequences exist in the literature. One is a process type of cognitive model including PARSER \cite{PERRUCHET1998246}, CCN \cite{anderson1990}, and others \cite{Rosenbaum1983HierarchicalSequences}. These models use heuristics to illustrate how chunks can arise from data, usually from jointly frequently occurring items that contain associative relationships. While PARSER compared simulated chunking with baby sequence segmentation when learning an artificial stream of language, CCN related the process of chunking with compositionality by organizing chunks in a hierarchical way. These process-level models are limited in their heuristics and lack a normative account of why chunking can be a rational behavior for the learning agents.  

In contrast to the process models, normative statistical models describe ideal observers’ behavior to explain why chunking is rational, usually using variants of the Bayesian ideal observer framework \cite{Goldwater2009}. For example, given a linguistic corpus, these models infer a segmentation with the highest probability from a set of chunks following the minimal description length principle. These models are rational as the inference is evaluated on observational instances. However, the inference process of these models suffers from combinatorial explosion with increasing sequence length. Presumably, the normative chunking models do not relate chunking with compositionality because of the computing complexity. 

Other chunk learning models are connectionist in nature. Approaches include using an artificial neural network to learn sequence segmentation or simulating spiking neural networks that are similar to our biological neural construct and translating the chunk learning problem into a loss function for network parameter optimization \cite{french_tracx_2011,cleeremans_finite_1989, Wang2017APlasticity}. Usually, these models generate behavioral consequences of chunk learning, but they are opaque to interpret due to their connectionist setup.  

This project implements computational cognitive models that connect chunk learning in cognitive science with the algorithmic advantages it provides. It also proposes a normative explanation, hypothesizing that chunking is a rational strategy. Building on previous experiments, which show that humans learn underlying chunks in sequences, this project suggests that humans behave like rational agents, uncovering patterns in their observations. Thus, chunking may serve as a rational method for identifying underlying relationships and regularities within observed sequences.


\section{Summary of the work}
To study the normative explanation of the chunking mechanism’s rationality while exploring the algorithmic advantage that such a mechanism brings forth, the project started by investigating the relation between hierarchical structure and chunking. To preciously control the structure of the sequence, we design a generative model that randomly comes up with an underlying nested hierarchical structure. The generative model starts with an inventory of initialized unique chunks of atomic units. In a number of iterations, existing chunks in the inventory randomly concatenate together to form chunk composites. Each chunk in the inventory is assigned an independent occurrence probability. The sequence is generated by consecutively sampling chunks from the generative model. This approach to developing a generative model captures the essence of compositionality at the sequence level: simpler concatenated chunks form foundational units that recur and combine to create increasingly complex structures within the hierarchy.

Observing such non-iid sequences with recurring chunks sampled from the hierarchical generative model, this project proposes that chunking can become a means to uncover the underlying structures in sequences. It proposes a simple chunk learning model that contains three components that can be plausibly implemented by cognition. 

The first component is parsing, i.e., the model parses and identifies chunks together as a unit from the sequences \cite{newellsimon1976,Gobet2001ChunkingLearning,perruchet2008learning,botvinick2012hierarchical}. The second component is learning the associative statistics of consecutively parsed chunks, a notion that inherits the legacy of behaviorists’ proposal (that animals learn to associate between events) while also being affirmed by the statistical learning literature (human learners are sensitive to the transition statics and the occurrence probability between consecutively observed entities in sequences \cite{gomez_artificial_1999, SAFFRAN1996, gomez_variability_2002,SAFFRAN199927}). Meanwhile, a forgetting component multiplies the count of parsed chunks by a discounting factor, a common practice that models forgetting \cite{ebbinghaus1913memory,murre2015replication,mozer2013reactivation,rabinovich2001dynamical,richards2017persistence,anderson2007how,Mozer2009}. 

Connecting the three components, this paper proposes the hierarchical chunking model, which builds up a nested hierarchical structure from sequences. HCM starts out learning about the minimally sufficient atomic sequential units as initial chunks to parse the sequence and combines chunks which has a correlated consecutive occurrence as indicated by the information provided by the associative statistics into more complex chunks to add to the dictionary. The simple merging process allows the model to learn more complex representations by reusing the previously learned chunks. In this way, a long and complex sequence can be learned as one entity in the dictionary by reusing and concatenating the existing chunks in the memory dictionary. The model learns a dynamical graph that is a trace of the evolving representation in the dictionary.

HCM brings the feature of compositionality in the sequence learning domain. As the previously learned chunks are used as basic components to parse the sequence and as candidates to compose into new chunks. The compositionality process contains a normative aspect guided by the recorded sequence parsing statistics, reducing the space of compositionality to those chunks that contain a correlated consecutive occurrence relationship and thereby also circumventing the vast search space as encountered by alternative formulations. Apart from that, the chunks become entities to factorize the high dimensional sequence distribution. One can evaluate the probability of a sequence $S =(x_1, x_2, \ldots, x_n) $ occurring $P(S) = P(x_1, x_2, \ldots, x_n)$ by the probability of chunks that constitute the sequence parsed by the model: $P(x_1,x_2)P(x_3,x_4)P(\ldots, x_n)$ (each group is the elements inside a chunk). Alternatively, the resulting representation can also be used as a generative model to produce imaginary sequences composed of sampling the learned chunks adhering to their occurrence probability by the model. 

HCM is formulated as a normative chunk discovery algorithm, i.e., chunking is rational for the model to uncover the underlying nested hierarchical structure in the sequence. This project includes learning guarantees of a rational HCM on an idealized generative model and demonstrated its convergence. Apart from formulating chunking as a normative representation discovery process, the project also shows several learning advantages this algorithm affords. 

The first one is data efficiency. On sequences with embedded hierarchies produced by the generative model, I compared HCM with RNNs learning from the same amount of sequential data. Given the same length of sequences, HCM could adaptively build its nested hierarchical representation by detecting correlation violations until no correlation can be detected. In contrast, neural networks are much slower at adapting their representation. Given the same amount of training data, HCM learned a better representation of the sequence than an RNN. We also observed that the advantage in HCM’s data efficiency becomes more pronounced as the hierarchy depth of the generative model increases. 

As HCM learns interpretable chunks, we also looked at the implication of transfer when the model adapts its previously learned representation to novel sequences generated by alternative hierarchical structures. The model's interpretability informs positive/negative transfer to learn representations in a novel environment where sequences come from a generative model with overlapping/complementary chunks. Since the previously learned chunks can be reused, the transparency of all existing chunks acquired by the model shows whether the new chunks need to be learned additionally. With full knowledge of the transfer sequences in relation to the learned representation from the model, positive or negative transfer can be reliably predicted. 

Many natural sequences may contain a hierarchical component similar to the generative model; as a testing ground, HCM was applied to learn structures from sequences from the book \textit{The Hunger Games}. From text sequences, HCM learns nested hierarchically embedded chunks reflecting the hierarchical organization structure of language. This includes the step-wise emergence of word parts such as common prefixes and suffixes in a word, commonly used verbs, and nouns, and later more complex phrases also emerge, including phrases such as “it is not just”, “in the school”, “our district,” and “cause of the”, similar to how we parse sentences through successive units of words and phrases when reading instead of letter-by-letter \cite{paterson2020}. 

This project also delves into the algorithmic consequence of chunking as a representation discovery mechanism in discovering interpretable compositional relationships from and beyond one-dimensional sequences, and into higher-dimensional visual and visual temporal sequences. I extended HCM to learn visual temporal chunks via proximal grouping and demonstrated that the model could learn frequently occurring visual-temporal parts to aid in breaking down a complex sequence of images into chunks of visual temporal wholes via combining their corresponding parts. Consequently, the complexity of the visual-temporal sequence reduces as learning progresses, and the model learns recurring visual-temporal movements. The model's behavior suggests that chunk learning can also capture the correlation in both spatial and temporal dimensions, hence may explain cognitive phenomenon beyond one-dimensional sequences to higher dimensions such as visual or proprioceptive sequences \cite{donderi2006visual}.

The ability to discover recurring temporal-spatial patterns and their sub-recurring patterns as chunks organized in a nested hierarchical graph makes HCM a method for extracting patterns in an unsupervised manner. One candidate data type hypothesized to contain a hierarchical structure is neural activities \cite{Alves2019}. We demonstrated that HCM can be used to learn recurring activations of functional brain regions on a resting-state fMRI data set. The interpretability of chunks allows matching the occurrence of chunks and stimulus onset to be compared with the known network connectivities in the brain. Via this method, we found nested network structures such as functional regions responsible for affect processing \cite{Stevens2011, WANG2015117}, visual attention control \cite{Vinette2015}, or theory of mind processes \cite{Bagozzi2013}. On a population level, we also observed a correlation between the average chunk size per participant and age - implying the aging brain possesses a more modularized activity signature. 

Together, we propose a model that processes a stream of perceptual sequence into chunks. HCM learns chunks as the basic unit of cognitive processing, allowing for the composition of chunks, factorization of sequences, and transferability to novel sequences. We demonstrate the application of this model to discover recurring patterns in an unsupervised manner, from one-dimensional sequences to multiple-dimensional visual-temporal sequences. This work suggests that chunking is a universal computational principle used to acquire parsable entities from sequences, resonating with previous discoveries in fields from sequence learning and memory to gestalt psychology and the arrival of visual entities. 

\section{Discussion}
\label{sec:srt:sec3}
The algorithmic design of HCM favors simplicity as a proof of concept to demonstrate the power of chunking. This simplification comes with limitations. One is the greediness in parsing; the model finds the biggest chunk in the learned dictionary in its volume to parse the sequence despite their low occurrence frequency as a heuristic. This choice may hinder the model from discovering the most plausible underlying chunks in the sequence, but rather in favor of expanding the inventory of its dictionary. Depending on the application, the future may extend the parsing process also to consider a model that infers the observation of chunk online \cite{Orban2008BayesianObservers}, thereby making the algorithm adhere to the updated probability of chunk parsing probability. 

Many pattern detection algorithms, including deep learning approaches, are not robust to independent noise in data. This poses big application problems and does not relate to the noisy biological substrate of the neural system, and hence is also difficult to implement in hardware systems that are also prone to noise or computation corruption \cite{goodfellow2015explaining,lecun2015deep,hardt2016train}.A feature of HCM is that chunk discovery is little affected by independent noise in sequences or when the occurrence instance of some underlying chunk is only partially observable by occasional signal corruption. This is because independent noise does not affect the statistical correlation among consecutively parsed chunks and, hence, does not influence the process of chunk proposal. What noise affects is the parsing process, i.e., smaller chunks consistent with the noise-corrupted sequences will be chosen as candidates for sequence parsing. As noise robustness is not the focus of this project, future work may exploit this feature advantage of the algorithm by simply making the parsing process more noise robust via introducing a similarity comparison mechanism matching the previously learned chunks and the sequence observed or adapting to some probabilistic variant, such as computing the chunk that leads to the maximal a posteri given a noisy observation sequence. 

Another limitation of this work is the computational efficiency in chunk searching. During each parsing step, the number of search steps to identify the biggest chunk that matches the sequence scales with the inventory size. Future work can improve the efficiency of this parsing step by either organizing chunks in a prefix tree-like structure to shorten the search step to scale with the depth of the prefix tree. Alternatively, chunk parsing can be implemented in parallel computing systems that contain independent components checking the consistency of individual chunks in the inventory simultaneously, with the biggest chunk matching in size winning the competition. Even better will be the combination of both types.

One innovation of this work is to relate chunking to compositionality in sequence learning, suggesting that chunking can be a means to compose the more complex from the simple in any problem domain that can be formulated as symbolic sequences. This formulation informs and differentiates from other approaches to model compositionality. 

The most prominent models that explain human compositional behavior are program induction models. They have been mostly applied to behavior domains with partial observability, such as explaining the composition of handwritten digits or coming up with programming steps that generate the output of a transformation from one base word form to its morphological variant \cite{Lake2015Human-levelInduction,ellis_dreamcoder_2020, ellis_synthesizing_2022}. 

Program induction models capture humans' capability to compose more complex cognitive representations from simpler ones by formulating mental computation as a combination of programs. 
Problem-solving involves searching for the combination of programs with the highest probability of explaining the observed data and updating its posterior distribution over the programs with more observation instances.

However, two problems hinder the program induction approach from being integrated to explain more human behavior or data domains with higher dimensionality. The first problem is that these models are domain-specific and assume knowing a set of computational primitives. However, these mental primitives are not known in most domains. Hence, such models have been limited to simple domains that are feasible for experts to hypothesize computational primitives, including hand-digit writing \cite{Lake2015Human-levelInduction}, geometric hand-drawing \cite{ellis_dreamcoder_2020}, or simple programming \cite{Rule2020}. The other problem is the vast program search space. Finding the right program combination that takes an input and produces an output is an exhaustive process and subject to combinatorial explosion. Improvements to program induction methods include using neural networks to guide the search process \cite{ellis_dreamcoder_2020}, but only partially alleviates this issue. 

By formulating learning in the sequential domain, our model of learning to compose entities to parse sequences captures the essence of compositionality while resolving both limitations of program induction approaches. Since the model starts learning with an empty inventory, primitives can be learned without the necessity to impose a library, and the model does not differentiate between the basic primitives or the frequently used composite chunks. Studying chunking in sequence learning also circumvents the huge computational complexity of finding combinations of primitive functions that explain an output observation, as combinations are acquired via observing sequences. 

While not every human behavioral aspect can be framed as inducing programs, most human behavior can be described by sequences of action and perception. Chunking, as a proposal for the emergence and reuse of subsequences as cognitive units, offers an explanation for a part of behavior in all domains that contain a sequential component. In future work, this approach can be integrated with the program induction methods by applying chunk learning to the sequences of program execution traces or other data domains where composing programs plays a role. 



Future work may also relate this model with cognitive experiments to test the implication of chunking during learning. If chunking as a tendency helps learning and reuse to acquire complex structures, then the curriculum should affect the learned representation critically. In particular, training needs to include enough repetition sequences to allow sufficient time for the learning agent to pick up the invariant patterns and rules. Indeed, studies have suggested that training regimes that fix invariant rules (blocked training) facilitate humans in learning about the underlying rules compared to regimes where the rule keeps changing (interleaved training) \cite{Flesch2018} - a feature that also distinguishes humans from artificial neural networks. Future studies can investigate the contribution of chunking to the curriculum effect observed in these experiments. 

Apart from the curriculum effect, the correlation detection feature of HCM can be applied to propose variables and generate hypothetical causal graphical models from sequential observations, subject to further model selections or interventions. 
Furthermore, applying chunking principles in the visual domain could explain the diverse Gestalt grouping laws. For instance, the tendency to group by similarity or common fate may be due to the correlated occurrence of similar objects in the visual field. Future studies may apply chunking principles to natural video data to test the hypothesis that some gestalt rules reflect the correlation of recurring patterns in natural image sequences. This could lead to the understanding that particular grouping principles result from an agent's identification of familiar visual relationships to reduce the perceptual complexity of observations, which is a rational strategy to break complex perception into parts.  


\section{Article Status}
“Learning Structure from the Ground up—Hierarchical Representation Learning by Chunking” (Wu, Elteto, Dasgupta, \& Schulz) \cite{wu_learning_2022} was published in \textit{Advances in Neural Information Processing Systems} 35, 36706 - 36721 (2022).

\section{Author Contributions}
\textbf{Conceptualization:} Shuchen Wu. \\
\textbf{Experiments:} Shuchen Wu, Ishita Dasgupta, Eric Schulz.\\
\textbf{Software:} Shuchen Wu.\\
\textbf{Visualization:} Shuchen Wu, Noémi Éltető.\\
\textbf{Writing – original draft:} Shuchen Wu, Eric Schulz.\\
\textbf{Writing – review \& editing:} Shuchen Wu, Noémi Éltető, Ishita Dasgupta, Eric Schulz.

\part{From concrete to abstract}\label{part:abstraction}

%
\chapter{Beyond concrete sequences --- Formulating and testing two types of motif learning in sequence learning and transfer} 
\label{sec:motif}
In the past chapters, we have investigated chunking experimentally and theoretically. From sequence learning experiments, we observed that humans adapt their chunk learning strategy to the underlying recurring chunks in sequences, suggesting that a normative pattern discovery principle can explain the chunk learning process. We delved into this question further in the second project, looking at sequences that contain a nested hierarchical structure. We proposed a model that adheres to the cognitive ability of humans to uncover chunks as entities of recurring patterns from sequences with a nested hierarchical structure. The model learns interpretable chunks that are also transferrable to novel sequences. 

Chunking has limitations in capturing the structural learning abilities of humans. It is a fascinating aspect of learning and cognition that we not only learn recurring patterns in their concrete form, but we are equally good at dismissing irrelevant details to learn abstract recurring patterns. Psychologists and neuroscientists related abstracting sensory experiences into concepts as preconditions of forming episodic memory during development. Concepts can be bound with context and chained into a memory of episodes, facilitating a recollection of concept sequences surrounding a personal event \cite{ghetti2011children,tulving1984precis}. Abstract concepts and categories become the seed of the thinking process, not only helpful for our memory but also helpful with logical induction and deductions of conclusions that lie outside of our personal experience  \cite{aristotle1938}. This chapter studies patterns in their abstract forms and their relation to transfer and generalization. 

Many daily examples suggest our instinctive attraction to motifs from sequences. Music, for instance, contains abundant melodic motifs invariant amongst varying note specifications. Beethoven’s Fifth immediately comes to mind when the iconic sequences of notes strike: GGGE, FFFD. Within the symphony, the note sequence progresses to GGGB or GGGC, with variations in forms and voices, one at each step.
The two examples point to two abstract sequence motif types that people are sensitive to, as the literature suggests. The first type relates to the initial definition of Gestalt (‘form’ in German) back in 1890: Von Ehrenfelds observed that a melody is recognized when played with different keys \cite{vonehrenfels1890ueber}. Later, this form of motif sensitivity was also suggested by the language learning literature. \cite{marcus_rule_1999} exposed seven-month-old infants to short sequences with simple grammar patterns such as ABA, QWQ, and EFE. This exposure made the infants sensitive to non-explicit sequence patterns, leading them to direct their gaze toward novel sequences sharing the same structure, such as KTK, rather than different structures, such as DDF. However, studies on these motifs usually look at the effect of motifs in very short sequences and do not describe how the abstract motifs may build up from the learning process. 

The second motif type, expecting that the progression of the music will vary on a position following the iconic three strikes GGG, relates closely to the linguist’s hypothesis on the acquisition of language, specifically grammar structure acquisition \citep{Boole1854, Marcus2001TheAM}. This type of motif is speculative to exist as a precondition of grammar structure acquisition. It also relates to Chomsky’s hypothesis that statistical learning between words cannot explain the infinity of language utterances --- a symbolic acquisition of language structure is necessary for people to judge the grammaticality of unseen sentences \citep{Chomsky1965, marcus1995, yang_universal_2004}. Learning abstract patterns at the symbolic level, such as the category of a noun, allows learning the abstract pattern of grammar, such as noun phrases typically consisting of a determiner followed by a noun. Developmental linguists suggest that children, after exposure to their native language, learn about the abstract category of nouns and verbs and are capable of applying their knowledge about nouns to novel phrases that have been seen to belong to the noun category. Sometimes, they overgeneralize the syntactical structure of nouns that demand an exceptional case \cite{Marcus2001TheAM}. This capability also demands the acquisition of language structure at the symbolic and abstract levels. However, such types of learning abstract structures in sequences have not been examined in an experimental setting. 

In this work, we expand on the previous literature with new additions. As the first kind of motif has only been tested in short sequences, we want to examine how people learn about these motifs in long sequences that are cognitively challenging. We look at the progress of how people acquire sequence motifs and how the two types of motifs affect the learning and memorization of novel sequences outside of participants' training experience.  

To do that, we first defined the two types of motifs in a sequence learning setting, especially the second type, as it has primarily been discussed in the language learning setting. We then formalize these motifs in a sequence learning setting to help study such motifs in a domain-general way. Following our definition, we conduct a sequence memory and recall experiment to test the effect of motif learning and motif transfer when people have been exposed to the two motif types and their ability to transfer to novel sequences that share the same motif type. The cognitively demanding task of remembering long sequences necessitates gradually building knowledge of the motif during the learning period.

Although literature have suggested that human cognitive capability is sensitive to abstract motifs in sequences, there have been no explanations for an underlying reason why people should learn motifs from sequences. Human motif learning has been discussed on an observational level, lacking a normative account. The modeling work we propose provides such a normative account: we suggest that learning abstract motifs can be closely related to learning chunks, and the process can be described by chunking on an abstract motif level. In this way, learning motifs can be explained by participants finding invariant structures in sequences for efficient compression. We build a model that integrates abstraction learning and chunking into the same program to discover sequence motifs manifested as abstract chunks in sequences.

\section{Toward a Taxonomy of Abstract Motifs} 
We define two types of motifs: projectional motifs and variable motifs.

A projectional motif is a pattern in a projected space shared among distinct sequences. A transformation function maps the superficial content to this projected space. For example, GGGE and FFFD's music phrases share the projectional motif XXXY.

A variable motif is a pattern with invariant and variant parts. In a sequence with a variable motif, a variable symbol represents a quantity that can change. These sequences share a structure with a varying entity at the “X” position and constant entities elsewhere. For instance, the music phrases GGGEZ, GGGB, and GGGC share a variable motif GGGX, with X taking the value of Z, or B, or C. 

\section{Summary of the article}
This article explores how abstraction aids in memorizing sequences and transferring abstract knowledge from one sequence to another in recall experiments. 

We tested this hypothesis in two serial recall experiments: participants were instructed to memorize 12 consecutively displayed colors and then recall the sequence by pressing corresponding keys, with recall accuracy recorded as the primary measure for analysis.

Experiment 1 examined how projectional motifs aid memorization and transfer. Sequences consisted of two variables, X and Y, each appearing 6 times per sequence presentation. Participants were divided into two motif groups (Motif 1; Motif 2) and a control group (Independent). A motif was consistent across training trials in the motif groups, while in the Independent group, X and Y were permuted in each trial. X and Y were mapped to distinct colors. Participants underwent 40 training trials followed by three transfer blocks, each consisting of 8 trials, testing motifs of each type. In the transfer blocks, sequence colors from training did not reappear. 
Experiment 2 tested the learning and transfer of variable motifs. Participants were divided into a variable motif group (motif) and a fixed group (control). The variable motif group memorized sequences like B X D F, with X varying (A, C, E), while the fixed group memorized sequences like B A D F, with no variation. Participants underwent 40 training trials followed by 24 transfer trials. In the test block, both groups memorized new sequences with variable X in the same positions as training but with changed fixed parts.
Analysis of the human recall accuracy data suggested that participants effectively learned and transferred both motif types. Training with variables and projectional motifs improved recall accuracy, especially on transfer sequences. 

We propose a model that differs one step from the hierarchical chunking model, integrating learning transition statistics, learning chunks, and pattern discovery on a motif level. To simulate sequence motif learning of the first type, we simulated the model by learning chunks in the abstract projectional motif space. For the motif of the second type, we integrated a component that proposes an abstract variable entity based on preadjacency and postadjaency transition statistics between the parsed chunks, thereby discovering recurring chunks in sequences that contain variables. Together, such a model simulates a progressive build-up of sequence motifs via discovering recurring patterns in the motif space and progressively concatenating the previously learned abstract chunks into bigger chunks, reusing the knowledge of sequence motifs to novel sequences. Simulation of the model in the two experiments learning identical sequential instructions to the participants suggested that the motif learning models progressively learn motif chunks, which help the model to transfer and generalize. The model sequence generation accuracy correlates with participants' sequence recall accuracy during the progression of the experiment. A detailed model comparison separately, including each component that consists of the motif learning model, suggested that motif learning and transfer cannot be explained by chunk learning or associative learning alone. Expanding chunking from concrete sequences to abstract representations was crucial for capturing the learning and transfer effects in this set of experiments.

Our findings suggest that human participants use both motifs to facilitate sequence memorization and generalization to novel, unseen sequences. This learning and transferring process in this experiment can be captured by chunk learning in the two types of abstract motif space. Discovering recurring patterns in sequences helps people memorize and transfer sequences with abstract motifs. Our work paves the way for a better understanding of how abstract motifs emerge progressively from sequences and their implications in generalization. 

\section{Discussion}
It is a fascinating aspect of learning and cognition that we not only learn recurring patterns in their concrete form, but we are equally good at dismissing irrelevant details to learn abstract patterns. 
Our work hypothesized two sequence motif types that humans could learn and generalize, tested these hypotheses in sequence memorization and recall experiments, and proposed a model that progressively builds up a complete sequence motif via chunking in abstract space. Our work advances our understanding of how people construct abstract representations from observational sequences for efficient compression and generalization. 

This work paves the way for future work to expand into the characteristics of learning abstractions in sequences. Our experiment tested abstract motif learning in a restricted number of sequence types: projection motif that spans the entire sequence and variable motifs that contain one variable at a specific ordinal position of the sequence; future work may expand upon the variability of this experimental paradigm and design experiments to study and test more flexible motif learning. For example, one can have, in the experimental sequence, multiple variable entities, X and Y, each having distinct entailment, located at different positions of the sequence, and look at how the learning of a sequence that contains Xs and Ys, helps to transfer to novel sequences, where the location of Xs and Ys may also swap. 

Additionally, future work may test hierarchies that span multiple abstraction layers, such as projectional motifs embedded in variable motifs, i.e., variables that represent several possible projectional motifs or vice versa, and how learning adapts to various motifs. Modeling-wise, this may correspond to the discovery of a abstract structure that helps a learning agent compress sequences based on the previously learned motifs. Future work can test the interaction between learning this motif structure and participants' performance in transfer and the individual variabilities in their sensitivities to either motif type in sequences.    

We studied motif learning in the sequence learning domain; future work may relate this work further to the general domain of learning abstractions. Many of these works on the role of abstraction and generalization formulate their problem in a non-sequential domain. For example, previous work on abstraction has studied our tendency to understand abstract concepts via metaphor, such as understanding the concept of an 'argument' in terms of 'war' and thereby transferring the feature of war to the concept of 'argument' \cite{LakeoffJohnson80}. Another example is in problem-solving, where people tend to find a solution to a new problem based on their knowledge of a familiar problem that resembles the new problem in abstract ways \cite{duncker_problem-solving_1945}. Additionally, acquiring reasoning rules from experience has been proposed to build the foundation for logical deduction and reasoning \cite{CHENG1985391}. Many of these works argue that finding commonalities among conceptual space manifested as abstract rules is fundamental to human intelligence. The property of abstraction has also helped to advance multiple fields. In math, abstraction empowers mapping deducted theorems from one axiomatic system to another \cite{millidge_towards_2021}. In computer science, abstracting computing steps into functions and classes allows the reduction of the computational complexity of programs \cite{abelson1996structure}. 

Future work may connect the sequential aspect of this model with the progressive acquisition of concept relational graphs or show how abstractions described by this abstraction literature can arrive from perceiving data sequentially. In particular, it can be interesting to adapt the model to describe a process of how abstraction structure can be built up progressively from learning: for example, the model can describe the process of realizing a solution to a simple problem via learning sequences that underlie the program traces of the search steps. In the meantime, arriving at sequences of simple abstract execution steps may help participants learn to piece together knowledge in the abstract space to reach higher-order abstraction manipulation. Models of such flavor can also be used to simulate and measure the learning difficulty of abstract problems, or the learning and developmental stages necessary as a precondition to understanding abstract concepts or transferring metaphorical understandings. Generally, the model may illuminate how simple mechanisms of chunking in an abstract space may help cognition find a common pattern beyond seemingly distinct observations and build up layers of cognitive sophistication to construct and extrapolate concepts outside our finite sequential experience. 

\section{Article Status}
“Motif Learning Facilitates Sequence Memorization and Generalization” (Wu, Thalmann, \& Schulz) has been submitted as a preprint on \textit{PsyArXiv} \cite{wu_thalmann_schulz_2023} and has been accepted in \textit{Nature Communications Psychology} doi:10.31234/osf.io/2a49z.

\section{Author Contributions}
\textbf{Conceptualization:} Shuchen Wu, Mirko Thalmann, Eric Schulz. \\
\textbf{Experiments:} Shuchen Wu, Mirko Thalmann, Eric Schulz.\\
\textbf{Software:} Shuchen Wu.\\
\textbf{Analysis:} Shuchen Wu, Mirko Thalmann.\\
\textbf{Writing – original draft:} Shuchen Wu.\\
\textbf{Writing – review \& editing:} Shuchen Wu, Mirko Thalmann, Eric Schulz.
%
\chapter{The construction from the simply abstract to the complexly abstract, layer by layer}
\label{sec:hvm}
\cleanchapterquote{Within great truth lies great simplicity.}{Lao Tzsu}{Tao Te Ching}

In the sequence recall experiments, we verified firsthand that two types of motifs help people memorize and transfer their knowledge to novel sequences. Our experiments suggested that people learn patterns not only on the sequence surface level but also on an abstract level. In the following work, I delve further into how chunking on an abstract level may help uncover layers of abstraction in data and how the mechanism of such a model relates to abstraction in general. 

The ability to form task-specific abstract representations has been suggested to be fundamental for our intelligence \cite{konidaris_necessity_2019,barsalou_perceptual_1999,piaget1954,dehaene_symbols_2022}. 
People are equipped with the ability to abstract. As we learn a new language, we also learn the salient patterns underlying grammatical forms without explicitly being told about the rules. For example, after learning German for a while, you will expect a verb at the end of a subordinate clause. This verb can mean “kick”, “support,” “drink,”... etc. But you develop a sensitivity to the functionality of the last word. Children acquire grammar structure when learning a language; they learn the rules, such as determiners precede nouns, and generalize the rules \cite{Dehaene2015TheTrees}. Infants as early as 23 months old can learn the category of nouns \cite{tomasello_twenty-three-month-old_1993,bernal2010two,melancon2015representations}, expect the syntactic category of the next word in a sentence, and use their knowledge about nouns in argumentative roles that they have not experienced in the past. 
Denoting unknown entities in symbolic abstract form was fundamental to the development of mathematics. It is easier to arrive at a solution of an algebraic equation such as “x + 5 = 10” by assuming ‘x’ as a symbolic, unknown entity to find out about “x = 5”. A proper abstract description may help an agent discover the underlying relation that governs the otherwise highly complex and variable observations. Consider Newton’s law in physics or Maxwell’s equation describing electromagnetic waves: abstracting unknown entities in symbolic forms has been civilization’s workforce to discover the invariant laws in nature.   

\section{Related work on modeling abstractions}
Knowing the theoretical and pragmatic implications of studying abstraction, researchers have attempted to model abstraction learning using different approaches. 

One approach to model abstraction builds explicit discrete conceptual relational systems manifested in graphical structures. This sort of model has been applied to explain human behavior, including understanding abstract concepts via linguistic metaphor \cite{LakeoffJohnson80}: such as grounding the concept of an ‘argument’ in terms of the definition of ‘war’ and thereby transferring the characteristics of war to the understanding of ‘argument’. 
Models with this flavor have also been applied to explain people’s transfer behavior in problem-solving, how solving a new problem becomes much easier when knowing the solution to a familiar problem that resembles the new problem on an abstract level \cite{duncker_problem-solving_1945}. These models usually represent knowledge or conceptual understanding in discrete forms, manifested in conceptual relational networks in which nodes are ideas or concepts, and edges denote the relation between the ideas. Modern adaptations of such approaches include the Probabilistic Analogical Mapping (PAM) model, which uses word embeddings created by neural network systems like BART (which maps concepts to vector embedding) to construct such a conceptual relational network \cite{ichien_predicting_2022}. The task of finding abstraction amongst the source and target analogical concepts or using the solution of a previous problem to solve a new problem can be translated into finding and applying graphical commonalities between the two discrete graphical structures \cite{lu_probabilistic_2021,saitta_model_2001}.
A limitation of this approach is that these models assume a conceptual relational structure or acquire them from connectionist systems and do not explain how learners build up the discrete conceptual relational graph from experience. 
In a similar vein, Kemp and Tenenbaum \cite{tenenbaum_how_2011} use a Hierarchical Bayesian model defined over a set of graph primitives and grammars to combine the primitive to illustrate how complex graphical structures can be acquired by combining simpler ones. However, the model relies on assuming a library of primitive abstraction relations and does not explain how the abstraction primitives may arise from data. Hence, these explicit discrete abstraction models have been primarily applied to restricted problems or data domains. 

Another approach to model abstraction circumvents the problem of finding an explicit representation via training connectionist neural networks through a variety of datasets that demand abstraction learning and transfer. Such approaches include meta-learning \cite{finn2017model,snell2017prototypical,gu2018meta}: training a neural network on a distribution of tasks that are in different domains but share some underlying properties \cite{binz2023meta,hospedales2021meta}. In these cases, neural networks can do one-shot or few-shot learning in novel tasks that share the underlying property with similar tasks in training data. These connectionist models contain implicit abstract representations in the learned weights \cite{WANG202190,yang2019task}, which are opaque to interpret what explicit rules or commonalities may have arisen from learning. Hence, understanding and interpreting these models is an area under active research \cite{yang2019task,driscoll2022representational,driscoll2020organizing,driscoll2023flexible}. Without knowing the explicit abstractions acquired by neural networks, it is even harder to compare with the type of abstraction and transfer ability that humans are using. Therefore, datasets or tasks that demand models to exert human-like abstractions and reasoning abilities are still challenging for the best connectionist models today \cite{chollet2019measure,mitchell2021ai,marcus2018deep}.

\section{The open question}
When studying the growth of abstraction from perceptual data in the cognitive system, it is crucial to develop computational principles that yield interpretable structured representations. These principles should be capable of learning structure from experience while maintaining interpretability. To achieve this, we can draw insights from the literature on developmental psychology. 

\section{How abstract concepts may grow inside the mind}
Literature suggests that abstract conceptual symbols originate in perceptual experience and arise from superficial sensory experiences. Indeed, evidence suggests a close relation between neural activities representing concepts and neural activities representing experiences. There are no specific neural substrates dedicated only to representing abstract concepts. Instead, during sensory-motor perceptual experiences, association areas in the brain capture bottom-up activation patterns in sensory-motor regions. Later, perceptual symbols activate the association areas, which in turn reactivate sensory-motor areas \cite{barsalou_perceptual_1999, glaser1992expert,cuccio_peircean_2018}.

The developmental literature suggests several key features of abstraction. 
The first feature abstraction is \textbf{commonality}. Classical conceptions suggest that abstraction arises from generalizing common features of experience. For example, the abstract concept of 'swans are white' arises after observing many instances of white swans \cite{yee_abstraction_2019}.

The second feature is \textbf{discretization}: a continuous information stream is divided into concrete, recurring, and symbolic units. There is an end, a beginning, and a range of values that an abstract concept assumes, such as the abstract concept of a noun includes cat, dog, box, and other words that belong to the noun category \cite{ohlsson_abstraction_1997}. 
The abstract categories that point to the words are articulated when parsing a string, such as \textit{The cat jumped out of the box} to check the grammatical validity. 

The third feature of abstraction is \textbf{information reduction}, i.e., throwing away information. An abstract concept is less specific than the concrete concept that it entails. The word \textit{cat} is more specific than the category \textit{noun}. Throwing away information has been suggested to be critical for people to learn higher-order statistical relationships that govern observation \cite{lynn_abstract_2020}. 

The fourth feature relates abstraction tightly to \textbf{generalization and transfer}. As in perceptual systems, there will never be an exact reoccurrence of the same data point. Abstracting from past experiences helps to develop concepts that can be reused to facilitate performance on a never-encountered task. 

Finally, abstract ideas can also be assembled. More complex abstract structures can be constructed via operating on existing abstractions, also referred to as coordination (of schemas) \cite{ohlsson_abstraction_1997, piaget1985equilibration}. 

While previous modeling work captures some of the abstraction features in developmental psychology, no model that captures all of them exists. This raises the question of what a minimal model can be that captures all of the abstraction features as described by psychologists. And what basic computational principle allows a learning agent to abstract while exhibiting the aforementioned features? Additionally, how can more complex, abstract structures be constructed by combining components from more superficial abstract structures in a way that is consistent and similar to the aforementioned developmental trajectories? 

I explore this question by controlling the generative model of sequences that favor abstraction and studying how the underlying structure can be uncovered by a learning agent. I propose to combine chunking and abstraction learning, being previously discussed in isolation, as the core mechanisms of this model. I argue that chunking --- in conjunction with learning abstractions --- can give rise to the ability to learn both concrete and abstract patterns while giving the power to assemble the complex from simpler parts. 

\section{Summary of the Article}

\subsection{Sequences with nested abstract hierarchical structures}
This project starts by studying one-dimensional discrete sequences as an extremely simplified version of perception. The perceptual sequence may reflect the underlying environment's inherent nested structures and regularities. To start with studying the necessity of abstraction, we develop a generative model to produce sequences that mimic the emergence of nested hierarchy in natural systems, taking references from the properties of self-diversifying systems \cite{von1999humboldt,abler_particulate_1989}. 

Specifically, this theory hypothesized that a set of simple principles constitutes the diverse observation in the natural world. Such systems ‘make infinite use of finite media’ whose ‘synthesis creates something not present in any of the associated constituents’. Within a self-diversifying system, a set of existing stable objects form stable combinations with one another to form more complex objects. This automatically leads to a variation and oversupply of created objects while producing stable combinations that share similar properties. Examples of such systems include the diverse chemicals constituted by atomic units, the diverse organisms constituted by a combination of genes, and the infinite possibilities constituted by finite means present in the human language. 

To capture this feature in the sequence subject to study, we simulated generative models that operate in one-dimensional sequences to simulate perceptual observations. The key assumptions of this model are:
\begin{itemize}
    \item All objects are made out of a finite combination of atomic elements. 
    \item The observation sequence is sampled from the created stable objects in the existing inventory of the world, where each object in the inventory occurs with a certain probability. 
    \item Existing objects may concatenate and combine into composites, forming new 'things' with similar properties that interact analogously.
    \item Some created objects share similar properties and belong to one category. This property will make them interact with other things in the world similarly, producing composites that only differ among objects within the same category.
\end{itemize}

The generative model begins with an inventory of basic atomic units. This inventory expands through the formation of novel stable combinations by concatenating existing objects or categories. Initially, atomic elements randomly combine to form objects, thereby expanding the inventory. Some of these objects become new categories, representing groups of objects that share similar interaction properties. These categories then serve as additional components, combining with other objects or categories to form new stable combinations to expand the inventory further. The agent observes random samples of objects from the inventory within the artificial world created by the generative model.

\subsection{Two ingredients of abstraction}
A learning agent perceives sequences that reflect the underlying nested structure created by the generative model and uses two types of abstraction, along with learning chunks, to uncover what are the unvarying patterns that occur in its perceptual stream. 

We introduce two implementations of abstraction notions in HVM that make the algorithm more effective while expanding its capability to uncover a hierarchy of variables. 

\textbf{Abstraction as organizing chunks via common subparts}
The first type of abstraction is finding common parts between the learned chunks. Upon parsing the observational sequence, the model needs to search among its existing learned chunks to retrieve one consistent with the sequence. The number of search steps to allocate the biggest matching item in the parsing tree grows with the size of the dictionary. 

Abstraction, as finding common parts between learned chunks, helps the model organize its memory more effectively for information retrieval. The memory of the learned chunks is implemented in a Trie structure: each ancestor node is the common prefix of its children, connecting the longer chunks with the shorter and more frequent chunks in a hierarchical memory recall graph. 
During parsing, the search starts from the root of the parsing tree, following each leaf node consistent with the sequence, and terminates at the deepest leaf node. Identifying the final node consistent with the upcoming part of the sequence is guaranteed to be the deepest chunk in the tree. Connecting chunks from their common prefixes reduces the search step to the depth of the tree. 

\textbf{Abstraction as inventing symbols that represent variables}
As perceptual sequences contain categories that similarly interact with other objects, uncovering these abstract concepts as categories helps the learning agent acquire higher-order patterns and relations that explain more observations. 

The second characteristic of abstraction is to replace the occurrence of distinct chunks using a symbol. The symbol is meant to denote categories of chunks sharing similar interaction properties. The symbol is identified when any chunk that this symbol represents is identified, which helps the agent to identify an underlying pattern in varying observations. This abstraction feature enables more abstract concepts to emerge from concrete patterns in a graded fashion, layer by layer, and the more abstract patterns detected based on the description of the previously acquired symbolic observation description, analogous to human abstraction concept formation during development.



We propose a hierarchical variable learning model (HVM) as an extension of the hierarchical chunking model that combines chunking with the two types of abstractions proposed above. HVM abstracts commonalities among its learned chunks to organize memory in a Trie structure to enhance retrieval efficiency. Furthermore, HVM proposes symbols to represent abstract categories of chunks with similar interaction properties. The model learns chunks on the description of previously learned symbolic patterns. This dual approach discovers abstractions by proposing variables to capture sequence variability and uses chunking operations on an expanding symbol inventory, mirroring concept discovery during cognitive development.

We first show that abstraction via extracting commonalities among chunks reduces parsing search steps. Additionally, proposing symbols to capture categories helps the model learn unvarying patterns that explain a larger part of the sequence. The models that exploit hierarchical structures compress the sequence more effectively than traditional compression methods.

Next, we showed the relation between compression efficiency, abstraction, and generalization. We showed that as the layer of abstraction increases, more abstract, symbolic chunks are learned by the hierarchical variable model, which comes with more distortion in pure symbolic representations and higher sequence parsing likelihood in novel transfer sequences. A more symbolic description of patterns in sequences helps the model to parse novel sequences with less surprise. 

Relating the model to human sequence learning, we used the same sequence recall experiment to instruct the model to remember sequences and compared the model's negative log-likelihood to participant sequence recall times. We discovered that the model's negative log likelihood correlates with human sequence recall more strongly than alternative models that do not learn and transfer chunks or variables. We further compared several large language models' (LLMs) negative log-likelihood in the memory experiment. In comparison to the cognitive models, we found that LLMs do not abstract.


\section{Discussion}
\label{sec:abs: conclusion}
In this work, we propose a model that learns chunks from the specific to the abstract levels. Chunking endows the learning agent with the ability to generate structural primitives as recurring patterns in sequences. Abstraction enables the agent to symbolize and organize akin patterns into categories. The interplay of the two forces enables learning unvarying patterns, from simple to complex, from concrete to abstract, growing with practice. The compositional nature of sequences encourages recursive reuse, building up the intricate wholes from intermediate parts. 

Our work goes beyond previous work in two aspects: the first is that assumptions on primitive abstraction functions are lifted and can be learned from data; the second is that an explicit abstraction can be acquired instead of relying on connectionist systems learning implicit abstractions. This work lays a foundation for a series of further investigations to elucidate the emergence of abstract structures from learning. 



Future work can relate this model to more cognitive phenomena or dive into the implementation level to understand the emergence of abstract structures in artificial/biological neural networks.  
One direction is to relate the parsing graph with memory retrieval. Using commonalities shared among memory items to organize memory implies that giving longer retrieval cues shall help allocate the retrieval and identification of a long memory faster and more accurately than shorter retrieval cues. Other experiments may relate the model's parsing steps with behavioral recognition time. A tree-structured parsing graph implies a logarithmic search time that grows with the number of stored memory items. Future work may test whether the memory retrieval time grows with the memory size or the logarithm of the memory size. 
Another direction is to relate chunking in the abstract space with the merge operation that allows a step-wise combination of corpus units to generate grammatically intact utterances \cite{Pinker1994,Chomsky1995,CHOMSKY201333}, due to their close resemblance.  
Additionally, the model predicts the existence of a particular type of memory error: items that appear in similar variable categories are likely to be confused during recall compared to memory items from different categories. Existing evidence suggests this to be the case \cite{carpenter2023within,gallo2003effects}. Further application of the model may include the explicit grounding of novel abstract concepts based on existing chunks to emulate our cognitive tendency to conceptualize the nonphysical in terms of the physical or the less clearly delineated in terms of the more clearly delineated. 

Abstraction also necessarily happens in large AI models used these days. Throwing away information and obtaining vital information is necessary to perform reasonably in any classification task and beyond. Previous work showed that the level of abstraction increases with the neural network's processing layer \cite{Kozma18}, and so does the level of information comprehension. However, how abstraction arises and its influence on transfer is still unknown, especially for modern AI systems. 
For substantially abstract tasks such as arithmetic or algorithmic binary operations, it has been observed that although some neural network models learn and predict very well on the training set after training, good performance on the test set does not emerge until training for an excessive amount \citep{power2022grokking, miller2024grokking}. Similarly, intriguing phenomena have been observed, such as a leap of reasoning ability emerging after excessive data \cite{wei2022emergent}. Other works on meta-learning suggested that training neural networks to perform multiple tasks helps the network to transfer skills and solve problems in novel situations. Our work demonstrates a tight relationship between abstraction and transfer. 
Our result differentiating models that learn chunks and abstractions on transfer sequences suggest that the acquisition of abstraction does not improve performance on training sets that representations on a superficial level can solve, but only on tasks that demand the usage of sufficiently abstract representation. LLMs' inability to exploit variable structure from the training sequence to the transfer sequences suggests the absence of a particular abstract representation that overlaps both the training and the test set. It relates to this question of how such an abstract structure may emerge at one point during the excessive training process. Our work urges future studies to study the emergence of abstraction at different learning stages by probing the model's behavior on tasks that demand different levels of abstraction and inform hypotheses such as are the lower, specific levels of abstraction necessary for the network to realize and come up with higher levels of abstraction in tolerance with higher variability in data. 

This algorithm can also be adapted for flexible applications. Future work can further improve computational efficiency adapting to the application in need. For example, one can implement a hash table on each branch of the chunk parsing graph to further reduce the computational steps of chunk parsing. Alternatively, the stored memory chunks can be organized via a semantic relational network or other structures to group similar memory items closer to each other in retrieval or storage space. In parallel computing systems, the time needed to retrieve and identify the correct stored chunk for parsing can be further reduced by harnessing neurally plausible such as a winner-take-all architecture to efficiently trigger the activation of neural populations representing the storage of chunks. 

Finally, the abstraction and chunking considered in the context of this work are in the domain of perception. Future work may integrate this modeling framework with action. Compositionality, transfer, and reuse models have also been proposed in classical hierarchical reinforcement learning and resemble human behavior \cite{xia_temporal_2020}. Abstraction in state space can alleviate the combinatorial explosion that plagues planning: by transforming the state space of a ground Markov Decision Process to that of an abstract one, task complexity can be reduced, paying a small loss of optimality. Approximate state abstraction condenses prohibitively large task representations into essential information and allows solutions to be tractably computable \cite{abel_near_nodate}. Future work could explore connections between this model and theories of hierarchical processing that integrate action and perception \cite{rao2024sensory, jiang2024dynamic, fisher2023recursive}. For instance, action could be incorporated as a mechanism to selectively focus on and verify information, aligning it with perceptual expectations.
 

\section{Article Status}
“Building, Reusing, and Generalizing Abstract Representations from Concrete Sequences” (Wu, Thalmann, Dayan, Akata, \& Schulz) is a submitted manuscript \citep{wu_thalmann_schulz_2024} and, at the time of submission, is under review at \textit{International Conference on Learning Representations}, doi:10.48550/arXiv/2410.21332.

\section{Author Contributions}
\textbf{Conceptualization:} Shuchen Wu. \\
\textbf{Experiments:} Shuchen Wu, Peter Dayan, Eric Schulz.\\
\textbf{Software:} Shuchen Wu.\\
\textbf{Analysis:} Shuchen Wu, Mirko Thalmann.\\
\textbf{Writing – original draft:} Shuchen Wu.\\
\textbf{Writing – review \& editing:} Shuchen Wu, Mirko Thalmann, Peter Dayan, Zeynep Akata, Eric Schulz. 
\part{Outlook}\label{part:outlook}


\chapter{Discussion}
\label{sec:discussion}
In this thesis, we have proposed computational models that learn concrete and abstract chunks from the ground up, uncovering and factorizing sequences with a nested hierarchical structure. The work also included human behavioral experiments, linking these computational models to human behavior. The models were applied to learn both abstract and concrete patterns in the general domain of sequence learning. Together, this thesis outlines a proposal for how structured representations may emerge from data, inspired by the cognitive mechanism of chunking. It also explored how previously learned chunks can facilitate the composition and reuse of knowledge, enabling the learning of more complex chunks at both concrete and abstract levels. 

Like all research, the work presented in this thesis has limitations. One key limitation of the proposed models is their reliance on a greedy parsing strategy. For simplicity, the models always select the longest chunk from the learned dictionary that aligns with the incoming sequence as the basis for parsing. While this heuristic encourages the learning of more complex chunks, it can also lead to rigidity. In some cases, selecting a longer chunk may not be the optimal choice if it has a lower likelihood of occurrence. This approach risks introducing dogmatism, where previously learned chunks overly influence how new chunks are discovered, limiting the model’s flexibility in learning alternative sequence fragments that could lead to a more diverse set of chunk entities.

Future work could, therefore, enhance both the rationality and flexibility of the parsing process. One potential improvement would be integrating sampling methods into the parsing algorithm, which could introduce more variability and adaptability in chunk discovery. Alternatively, the parsing strategy could be adapted to account for partial observability in sequences where chunks might be incompletely visible. This could be achieved by sampling the chunk that most closely aligns with the observed sequence based on both prior knowledge and their occurrence likelihood. Incorporating Bayesian inference into the parsing process would allow the model to infer the most likely chunks given the sequences observed so far, improving the model’s capacity to learn more flexible and diverse dictionaries.

Another limitation of this work lies in the potential mismatch between our generative model and the perceptual reality we aim to describe. In this thesis, we related chunk learning to a rational process of discovering underlying patterns in sequences, leading us to propose a generative model that produces sequences with a nested hierarchical structure. The chunk-learning models we developed served as recognition models that approximate the inverse of this postulated generative process. However, the assumption of a hierarchical structure may not hold in all domains of sequential data. Some sequences might have an entirely different structure or lack any hierarchy, such as patterns that do not exhibit spatial or temporal continuity.

In such cases, the chunk-learning models proposed here may generate an excessive number of chunks, diverging from an optimally succinct representation of the underlying sequence. For example, consider a sequence where each number is always a multiple of the previous one — neither the HVM nor HCM models would capture this particular rule. Thus, in domains where the underlying structure deviates significantly from the hierarchical assumption, the models may fail to learn useful data representations. In this case, the data would need to be decomposed or transformed into an alternative space that contains a recurring structure to enjoy the advantage of this type of model. This potential deviation highlights the need for future research to explore sequence structures and quantify such mismatches. Since the true generative processes behind sensory data are unknown, studies should aim to align generative models with the specific types of sequences being analyzed. Different real-world sequences may have distinct structural properties. Future work could bridge the gap between the generative model and actual sequence structures by identifying statistical properties, such as power-law coefficients \cite{piantadosi2014zipf, ferrer2001theory}, or using alternative measures of compositionality \cite{moran2013complexity, manning1999foundations} to characterize the deviation between the generative model and the structure of data at hand. In sequences generated by processes that do not guarantee spatial or temporal continuity, it could be interesting to test if the tendency to chunk may mislead humans to learn inefficient or faulty patterns, a seemingly irrational behavior caused by a system evolutionarily adapted to a particular data type.

Despite the limitations in some domains, this hierarchical assumption may underlie many sequential data types, including language or visual-temporal sequences. Applying chunking models to such data can provide valuable insights for practitioners looking to extract structured representations. One particularly exciting direction is behavioral data. From nematodes to fruit flies, from mice to humans, neural ethologists have observed that animals exert complex behavioral repertoires by recursively combining behavioral movement primitives, and has long been postulated that behavior is fundamentally organized by a hierarchical structure \cite{Baerends1976TheFO, Tinbergen1951, Berman2016,stephens2008dimensionality,wiltschko2015mapping,miller1950plans}. However, this hypothesis has been difficult to test on movement recordings due to a lack of methods to extract behavioral hierarchy in an unsupervised manner \cite{Berman2016,JOHNSON202070,DATTA201911}. Future work could apply and extend the models proposed in this thesis to automatically extract animals' “behavioral syllables,” allowing an automatic decomposition of complex behavior into modularized hierarchical structures. Decomposing behavior into hierarchies will allow future scientists to study and test hypotheses that have been primarily approached in a qualitative way, bringing an understanding of movement organization into a quantitative domain. Example questions include how behavioral dictionaries are constructed, the organization of movement motifs, how animals chain these motifs to form complex sequences, and how the psychological or energetic state influences the composition of movement motifs. Relating the movement motifs to animal's neural activities may also provide insights into the neural basis of sequence chunking. 


Beyond the limitations of the greedy algorithm and assumptions about the generative model, this thesis addresses chunk learning at the computational and algorithmic levels of Marr’s \cite{marr1982vision}. It explores chunking from a computational principle point of view and proposes that the goal of chunking is to find what are the underlying invariant entities in observational sequences. The thesis then proposes an algorithm with minimal but cognitively plausible components to learn chunks. Future work shall build on top of this framework and study chunking on Marr’s implementation level — how chunk learning can be implemented by a neural system within a biological substrate. This could involve exploring the biological mechanisms that give rise to sequence chunking and hierarchical compositionality in behavior, and identifying neural interactions that might give rise to equivalent computation of associative and chunk learning as included by the cognitive models in this thesis.

In the past, I have explored this implementation-level question by asking how chunking can be implemented in neuromorphic circuits. In mixed-signal neuromorphic hardware that emulates the firing activities of simple biological neural circuits, we have demonstrated a group of spiking neurons with synaptic plasticity and homeostasis can efficiently parse chunks and learn nested patterns from sequences. This work suggests that the parallel computation of the neural system is naturally efficient for learning and retrieving a successive activation of neural sequences in a computationally efficient way \cite{schreiber2023biologicallyplausible}. Additionally, chunk learning can be an algorithm that is especially efficient for a parallel computational system to learn structure via interacting with the environment, a property that the brain exhibits. 

Another way to investigate this implementation-level question is to study the neural correlates of chunk learning directly. Literature suggests neural substrates of chunk learning in the human brain, such as neural oscillation frequencies, reflecting the nested structure in linguistic sequences on an organization level of syllables, words, and phrases \cite{kaufeld_linguistic_2020}. These findings resonate with our discovery of rich nested structures in fMRI data \cite{wu_learning_2022}. Future work may adapt this model to identify groups of coordinating functional brain regions or neural population activities while participants are listening to or reading linguistic sequences and use this method to look at how the neural circuits can be held accountable for the emergence of pattern identifications across a variety of linguistic organizational levels. 

Alternatively, future work may also explore the neural basis of recurring action sequence patterns directly in biological neural activities. Previous research has observed transient, sequential neural firing activities across species and multiple brain areas, which have been linked to cognitive processes including animal spatial navigation \cite{vaz_replay_2020}, learning \cite{Foster2006}, sleep \cite{Lee2002, Wilson1994}, planning \cite{Diba2007}, and the encoding and switching of abstract rules \cite{wallis_single_2001}. These firing sequences often exhibit a hierarchical structure \cite{Bressler2006, Harris2005,Luczak2015}. In humans, similar sequences of neural activity have been implicated in cognitive operations like memory retrieval \cite{vaz_replay_2020}, consolidation \cite{Gelbard-Sagiv2008, Carr2011, Eichenbaum2013}, planning \cite{vaz_replay_2020}, and creative thought \cite{buzsaki_hippocampal_2015, joo_hippocampal_2018, buzsaki2010neural}.

The models presented in this thesis could be adapted to identify and separate recurring patterns of transient neural firing among neural population recordings. These neural sequences could then be interpreted as entities that correspond to specific behavioral patterns. Advancing in this direction could lead to a deeper understanding of how biological neural systems learn, adapt to, and process recurring patterns in sensory sequences. Additionally, hypotheses such as whether the hierarchical organization of behavior is driven by a hierarchical organization of neural activities, which may originate from the structure of naturalistic data, could be tested. This knowledge could, for example, inform the development of improved learning curricula. 


Finally, the algorithm proposed here operates primarily on one-dimensional discrete sequences. HCM is extended to learn chunks in higher-dimensional data but limited to 625 dimensions; future work can extend the algorithm further to learn structured representations in higher-dimensional sequential data, such as visual-temporal, proprioceptive, or frequency/sound domain.

Taking sequential data in the visual-temporal domain, for example, symbolic computer vision models have seen the visual perception process as an inverse recognition process to discover unvarying entities in the visual data. Exemplified by works by Zhu et al., computer vision researchers have tried to come up with a set of image primitives and image grammar in order to parse objects, scenes, and events as entities that are interpretable and robust under occlusion and signal perturbation \cite{zhu2021cognitive,zhu2021statistical,zhu2021stochastic}.

Although limited in its scalability, this previous approach informs future work to integrate chunk learning with connectionist systems to arrive at structured parsing of image/video data. Instead of learning directly from the full image space, which usually contains high dimensionality and variability, future work can take the embedding layer of pre-trained neural network models on vision data such as visual transformers and their variants \cite{wang2021pyramid,carion2020detr,liu2021swin,touvron2021deit,dosovitskiy2020vit,Vaswani2017AttentionNeed}, and directly look for recurring entities in the embedding layers, as training on a downstream task shall force the connectionist system to learn compressed representations. Other dimensionality reduction methods, such as vector quantized variational autoencoders, may also facilitate the reduction of embedding space to a manageable lower dimensional space. 
Applying chunking models to a low-dimensional embedding space or subsequent processing by some similarity-matching algorithms could extract interpretable symbolic entities from implicit intermediate network layers and reveal recurring entities in neural activity patterns, potentially corresponding to recurring entities in the input data. These entities could be perturbed to influence the behavior of downstream neural networks. Furthermore, manipulating this structured representation, such as combining these chunks, may create an explicit compositional structure within the embedding space, which could help to disentangle factors that independently influence observation data, allowing the downstream decoder network to generate data that adheres to the compositional structure. This approach might also offer insights into the texture bias in computer vision models compared to the shape bias in human vision, possibly due to the different chunk content learned by these two systems \cite{Geirhos2018GeneralisationNetworks}.

More significantly, applying variants of HCM and HVM to the embedding spaces of connectionist models that process high-dimensional data could bridge the gap between discrete entities identified by perception and the high-dimensional continuous representations typical of connectionist systems. This may enable models to form object-based representations and learn relations defined at a symbolic level, thereby improving the reliability of connectionist models while enabling the learning and transfer of previously learned entities, potentially reconciling the gap between symbolic and connectionist approaches to artificial intelligence and arriving at a modern solution to an ancient problem. 

\chapter{Conclusion}
\label{sec:conclusion}


Navigating a bustling city may seem effortless, but beneath this fluid interaction lies a remarkable cognitive feat: the ability to continuously process a torrent of sensory data, distinguishing relevant entities—such as traffic signals, road conditions, and food stands—from the chaotic stream of perception. This seamless parsing of the world into discrete units is fundamental to human cognition, yet both connectionist and symbolic AI models have struggled to explain how entities emerge from perception in such a natural, efficient manner. For centuries, philosophers, psychologists, and cognitive scientists have debated this core process, and this thesis focuses on chunking as a plausible cognitive mechanism that bridges this gap.

The thesis begins by experimentally probing human chunking behavior in a serial reaction time task. The results of two experiments suggest that people adapt their chunking strategies dynamically, attuned to both the statistical properties of sequences and the demands of the task at hand. The observed behaviors align with a rational model of chunk learning that strategically balances the trade-off between speed and accuracy—a principled computational account of efficient segmentation and organization.

Building on these insights, the second project explores chunking from a computational and normative standpoint. A generative model is proposed that unearths hierarchical patterns within sequences, recursively uncovering nested structures that serve as the fundamental building blocks of perception. These chunks become entities, not merely for understanding one-dimensional sequences, but as reusable and composable components in more complex, multidimensional environments. The proposed model is extended to learn part-whole structures from visual-temporal data, uncovering meaningful patterns in brain function and demonstrating the versatility of chunking as a learning mechanism.

In the third project, chunking moves beyond concrete sequence elements to capture more abstract motifs. Participants in two additional experiments progressively learn and transfer these motifs across cognitively demanding serial recall tasks, suggesting a deep, transferable chunking process. The corresponding model captures similar transfer behaviors, distinguishing between superficial chunk associations and deeper, abstract motifs. This reveals chunking's potential as a mechanism for abstract pattern recognition, with implications for how humans generalize from past experiences to novel situations.

The final project pushes the boundaries of chunking further into the realm of abstraction, proposing a hierarchical variable model (HVM) that layers abstraction over concrete chunking. This model not only compresses and organizes information through chunking but also generalizes these chunks into abstract variables, yielding a structured, flexible memory organization. The model’s striking alignment with human behavior demonstrates its potential to outperform large language models by learning contextually relevant, nested categories, emphasizing both memory efficiency and generalization.

Altogether, this thesis argues that chunking—a simple computational mechanism for associating nearby chunks in temporal and spatial proximity—serves as a fundamental process for singling out stable entities from the noisy perceptual stream. Through computational modeling and empirical evidence, the thesis shows how chunking, starting from scratch, can scaffold an agent’s understanding of the world, forming the basis for a structured world model learned through experience.

This work advances our understanding of chunking beyond traditional sequence learning. It illuminates how chunking compresses and organizes perceptual data, supports the learning of part-whole hierarchies, and facilitates compositionality and transfer across abstract domains. The thesis underscores chunking’s centrality in human cognition, intersecting areas such as associative learning, Gestalt theory, and grammar acquisition. It calls for future investigations into chunking's role as a discovery mechanism across cognitive domains, inviting further exploration of its neural basis, behavioral relevance, and potential as a method for uncovering hierarchical regularities in complex data.

\cleardoublepage

{%
\setstretch{1.1}
\renewcommand{\bibfont}{\normalfont\small}
\setlength{\biblabelsep}{0pt}
\setlength{\bibitemsep}{0.5\baselineskip plus 0.5\baselineskip}
\printbibliography
}



\cleardoublepage
\pdfbookmark[0]{Contribution}{Contribution}
\chapter{Statement of Contribution}
\label{sec:contribution}
\thispagestyle{empty}
\textit{This chapter encloses a statement of contributions provided to abide by the guidelines of the Graduate Training Center.}

The work reported in this thesis is entirely my own. All chapters have been directly supervised by my primary supervisor, Prof. Dr. Eric Schulz.

\textrm{\textsc{Chunking as a rational solution to the speed-accuracy trade-off in a serial reaction time task}}

\textit{Framework:}
I authored this paper as the lead author as the first individual project of my PhD. This project was conducted under the main supervision of
Prof. Dr. Eric Schulz (ES) and in collaboration with Noemi Éltető (NE) and Dr. Ishita Dasgupta (ID). I am the sole first author on this paper.

\textit{Contributions:}
\begin{itemize}
    \item I designed the experiments under the supervision of ES and suggestions from NE and ID. It was during the initiation phase of COVID-19 that we experienced some delays in obtaining ethical approvals.  
    \item I programmed the experiments as an online study, piloted it and collected the data. 
    \item I developed the mathematical models, implemented them in code and ran and analyzed simulations under the main supervision of ES. ES helped with the data analysis for experiment 2. ID and NE provided suggestions for data analysis.
    \item I wrote the entire initial paper and revision under the main supervision of ES. ES, ID, and NE provided suggestions and editing.
\end{itemize}

\textrm{\textsc{Learning Structure from the Ground up — Hierarchical Representation Learning by Chunking}}

\textit{Framework:}
This paper was the second individual project of my PhD. I am the sole first author on this paper. This project was conducted under the main supervision of
Prof. Dr. Eric Schulz (ES) and in collaboration with Noemi Éltető (NE) and Dr. Ishita Dasgupta (ID). 

\textit{Contributions:}
\begin{itemize}
    \item I came up with the conceptualization of the paper, developed the model, and programmed the model and its generalization to higher dimensional sequences. 
    \item NE and ID provided suggestions for the model comparison experiments. NE helped with the model comparison with PARSER and AL. 
    \item ID suggested the transfer experiments. 
    \item ES suggested the experiments on fMRI data. 
    \item I wrote the entire initial paper under the main supervision of ES. ID, and NE provided suggestions and editing.
    \item I discussed extensively with ES to come up with experiments and analysis during revision. I programmed the experiments and evaluation. ID and NE provided additional suggestions for editing.
\end{itemize}

\textrm{\textsc{Motif Learning Facilitates Sequence Memorization and Generalization}}

\textit{Framework:}
This paper was the third individual project of my PhD. I am the sole first author of this paper. This project was conducted under the main supervision of
Prof. Dr. Eric Schulz (ES) and in collaboration with Dr. Mirko Thalmann (MT). 

\textit{Contributions:}
\begin{itemize}
    \item I conceptualized the paper together with ES. 
    \item I developed the model and programmed the model. 
    \item I designed the experiments together with MT and ES. I programmed the experiments and conducted iterations of pilot studies with suggestions from MT and ES.   
    \item I analyzed the data. MT and ES provided suggestions on data analysis.  
    \item I wrote the entire initial paper, MT provided extensive editing, which went on several iterations with ES. 
    \item I discussed extensively with ES and MT on additional analysis. I conducted additional analysis during the rebuttal. MT and ES provided suggestions and editing.
\end{itemize}

\textrm{\textsc{Building, Reusing, and Generalizing Abstract Representations from Concrete
Sequences}}

\textit{Framework:}
This paper was the fourth individual project of my PhD. I am the sole first author on this paper. This project was conducted under the supervision of
Prof. Dr. Eric Schulz (ES) and in collaboration with Dr. Mirko Thalmann, Prof. Dr. Peter Dayan (PD) and Prof. Dr. Zeynep Akata (ZA). 

\textit{Contributions:}
\begin{itemize}
    \item I conceptualized the paper, developed the algorithm and evaluation methods and programmed the model and experiments. 
    \item PD suggested the comparison and discussion in relation to compression and models of compression. 
    \item ES suggested the LLM experiments and the evaluation of language sequences. 
    \item I wrote the entire initial paper. PD, and ES provided extensive suggestions and editing. MT, and ZA provided additional suggestions and editing.
\end{itemize}

\cleardoublepage
  This appendix includes the original papers and preprints discussed in this thesis.

  \begin{enumerate}
    \item Chunking as a rational solution to the speed–accuracy trade-off in a serial reaction time task (Wu, Élteto, Dasgupta, \& Schulz, \textit{Nature Scientific Reports} 13, 7680 (2023), doi:10.1038/s41598-023-31500-3)
    \item Learning Structure from the Ground-up—Hierarchical Representation Learning by Chunking  (Wu, Élteto, Dasgupta, \& Schulz, 2022, \textit{36th Conference on Neural Information Processing Systems (NeurIPS 2022)} 35, 36706 - 36721 (2022)). 
    \item Two Types of Motifs Enhance the Recall and Generalization of Long Sequences. Preprint  (Wu, Thalmann, Schulz, 2023, \textit{PsyArXiv} and has been accepted in \textit{Nature Communications Psychology}, doi:10.31234/osf.io/2a49z)
    \item Building, Reusing, and Generalizing Abstract Representations from Concrete Sequences (Wu, Thalmann, Dayan, Akata, \& Schulz, 2024, is under review at \textit{International Conference on Learning Representations}, doi:10.48550/arXiv/2410.21332)
  \end{enumerate}

\cleardoublepage
\part{Manuscripts}\label{part:manuscripts}




\phantomsection
\chapter*{Chunking as a rational solution to the speed-accuracy trade-off in a serial reaction time task} 
\addcontentsline{toc}{chapter}{Chunking as a rational solution to the speed–accuracy trade-off in a serial reaction time task}\label{pap:srt}
\includepdf[pages={1-}]{manuscripts/wu2021.pdf}

\chapter*{Learning Structure from the Ground-up—Hierarchical Representation Learning by Chunking
}
\addcontentsline{toc}{chapter}{Learning Structure from the Ground-up—Hierarchical Representation Learning by Chunking}\label{pap:hcm}
\includepdf[pages={1-}]{manuscripts/wu2022.pdf}

\chapter*{Two types of motifs enhance the recall and generalization of long sequences}\label{pap:motif}
\addcontentsline{toc}{chapter}{Motif Learning Facilitates Sequence Memorization and Generalization}
\includepdf[pages={1-}]{manuscripts/wu2023.pdf}

\chapter*{Building, Reusing, and Generalizing Abstract Representations from Concrete Sequences
}\label{pap:hvm}
\addcontentsline{toc}{chapter}{Building, Reusing, and Generalizing Abstract Representations from Concrete Sequences}
\includepdf[pages={1-}]{manuscripts/wu2024.pdf}

\cleardoublepage




\chapter{Afterward: From Dionysius emerges Apollo}
\label{sec:concepts}
\cleanchapterquote{Wherever the Dionysian prevailed, the Apollonian was checked and destroyed.... wherever the first Dionysian onslaught was successfully withstood, the authority and majesty of the Delphic god Apollo exhibited itself as more rigid and menacing than ever.}{Friedrich Nietzsche}{The Birth of Tragedy}

At the beginning, there was only the wild, untamed world of Dionysius. 
The god of wine's spirit ran wild like feral vines, thick and twisting, spreading across the land. Sensations, emotions, and desires flowed freely; chaos reigned supreme. Every moment was saturated with stimuli --- flashes of ecstatic joy, rumbles of anger, and swells of passion. Dionysius' followers, ecstatic and fervent, sang in drunken unison, their feet pounding the earth in a frenzied ritual to honor their god. Everything is connected but without clear distinction ---raw, intense, and boundless. The world was too vast, too complex to comprehend all at once, threatening to overwhelm the gods and mortals alike.

As Dionysius continued his wild revelry, a seed force began to stir deep within the heart of this madness and sensory overload. It was a faint glimmer of form within the formless, a rhythm within the dissonance. This seed was Apollo. 
Apollo noticed the repetition in the madness, the patterns beneath the confusion. The dancers, though frenzied, reiterated certain steps. The vines, though wild, grew in predictable directions. Even the songs of the revelers, returned to familiar melodies.
According to the noticeable patterns, Apollo \emph{chunked} the sequences of chaotic sensations into manageable, meaningful units: the once incomprehensible swirl of sensation was sorted into clear entities, each with its own beginning, middle, and end. The frenzied dancers slowed as their movements took on the grace of choreographed steps. The wild music began to follow a rhythm, its notes falling into place. The dancing bacchanals became structured rituals, their movements still passionate but now guided by purpose and harmony. The wild screams turned to song, each note precise, each rhythm intentional.

Through the process of chunking, Apollo — the god of sunlight, music, and prophecy — emerged as a radiant presence, taming the chaos of perception. He transformed Dionysius' boundless primal power and wild energy into something finite and comprehensible, distilling raw, unfiltered experience into meaningful chunks and symbols. In doing so, laying the foundation for reason, knowledge,  art, and beyond. Together, the two gods existed within the same perceptual world while governing a dichotomous characterization of reality: Dionysius was the intoxication of raw experience — perception as disordered and undifferentiated forms; Apollo was the clarity that followed — the ordered understanding that emerged after the chaos, differentiated by forms. 


\mbox{}

\end{document}